\pdfoutput=1
\documentclass[10pt,twocolumn,letterpaper]{article}

\PassOptionsToPackage{table,xcdraw}{xcolor}
\usepackage{wacv}
\usepackage{amsmath,amssymb}
\usepackage{booktabs}

\definecolor{wacvblue}{rgb}{0.21,0.49,0.74}
\usepackage[pagebackref,breaklinks,colorlinks,allcolors=wacvblue]{hyperref}

\def\wacvPaperID{24}
\def\confName{WACV}
\def\confYear{2027}

\title{Anatomy-Aware Pseudo-Labels and Homotopy-Annealed Hausdorff Supervision for Semi-Supervised Pancreas Segmentation}

\author{
Alireza Kheyrkhah$^{1}$,
Amir Hossein Saleknia$^{1}$,
Reza Azad$^{2}$\\
Sina Houshmand$^{3}$,
Dorit Merhof$^{2}$,
Ulas Bagci$^{4}$,
and Alaa Sulaiman$^{5}$\\[0.6em]
$^{1}$Independent Researcher,
$^{2}$University of Regensburg, Germany\\
$^{3}$University of California San Francisco, USA,
$^{4}$Northwestern University, USA,
$^{5}$Abdora Inc., USA\\
{\tt\small alaa@abdora.ai}
}

\begin{document}
\maketitle
\begin{abstract}
Confidence-gated co-training assumes that pseudo-label errors are uncertain
and concentrated at organ boundaries. We argue that for pancreas CT, where
the target organ is a single connected structure and detached fragments are
anatomically implausible, the dominant failures arise at three levels:
\emph{where} the organ is, \emph{what} its boundary should look like, and
\emph{how} a correct boundary is penalized. Anatomical Rectification relabels
implausible fragments of a teacher prediction as confident background,
without adding any loss or parameter; an
audit confirms that its rejections are overwhelmingly detached
hallucinations and that thin, tail-like slices are not eroded. Boundary
Grafting transplants labelled surface regions with their annotation
into unlabeled volumes, replacing unreliable boundary pseudo-labels with
ground truth. The residual error is coherent surface displacement, weakly
penalized by permutation-invariant Dice and cross-entropy losses.
Homotopy-Annealed Hausdorff (HAH) supervision anneals a
distance-weighted penalty from
an annotation-anchored objective that is convex in the prediction to a
Hausdorff surrogate, admitting the prediction-dependent distance term only
after the predicted surface becomes reliable. On NIH pancreas CT, under the standard semi-supervised split, the full method reaches
$84.60\%$ Dice and $4.97$ voxels 95HD, improving all four metrics over compared methods; on MSD Task07, $78.90\%$ Dice and $4.19$ voxels 95HD. We further contribute PanTS-SSL, a
semi-supervised evaluation protocol derived from the public 9{,}901-case
PanTS corpus, with a curated 1{,}015-volume main pool, labelled tiers of 50,
100 and 250 volumes, an 80-case tumor track, shift probes for acquisition and
anatomy change, and a fixed 901-volume test set. On it, the method exceeds the co-training baseline at 100 labels, scales to 85.29\% Dice at 250 labels, and reaches 85.50\% Dice at 100 labels with HAH.
\end{abstract}

\section{Introduction}
\label{sec:intro}

Voxel-level annotation of abdominal CT is costly, and the pancreas is among the hardest organs to delineate: it tapers from centimetres at the head to millimetres at the tail, borders fat and bowel at low contrast, and varies across patients \cite{roth2015deeporgan}. Semi-supervised learning answers with consistency: a network agrees with its own perturbed or temporally averaged predictions \cite{tarvainen2017mean}, or two independently initialized networks supervise each other \cite{blum1998combining}, keeping their trajectories distinct and avoiding the confirmation bias of a single teacher-student pair \cite{berthelot2019mixmatch}. We adopt translation-consistent co-training \cite{liu2024translation}, a strong instance of this family on pancreas CT, unmodified as our starting point.

These recipes share one instrument, a confidence gate that decides which pseudo-labels enter training \cite{sohn2020fixmatch}, sometimes sharpened by uncertainty \cite{yu2019uncertainty}, and one tacit assumption: the errors it must catch are uncertain and concentrated at organ boundaries. For the pancreas this assumption is misplaced, for anatomical reasons: the pancreas is a single connected structure, so a prediction splitting into disjoint fragments is impossible however confident, yet a confidence gate is blind to this because it reads each voxel only through the peer's confidence. We therefore hypothesize that confidence-gated co-training fails at three levels the gate cannot see: \emph{where} the organ is, since confidently detached fragments are reinforced throughout training; \emph{what} its boundary should look like, since the two networks converge on shared boundary mistakes that no voxel-wise rule can separate; and \emph{how} a correct boundary is penalized, since permutation-invariant losses such as Dice and cross-entropy charge a boundary displaced by a few voxels no more than an interior flip. The admission statistics and the rejection audit supporting this account are in Secs.~\ref{sec:method} and~\ref{sec:ablations}.

Each level suggests a correction; the first two act on the pseudo-label distribution, not on the objective. Anatomical Rectification (AR) supplies the missing criterion: fragments small relative to the main component of a teacher prediction are relabeled as confident background, telling the student where the organ cannot be rather than merely withholding the voxel, in the spirit of negative learning \cite{ishida2017learning}; when the prediction is already connected the edit vanishes. Boundary Grafting (G) supplies the missing targets: a targeted fraction of admitted boundary supervision is wrong and correlated across the two networks, so no reweighting recovers it; G instead transplants a cuboid sampled on the surface of a labelled volume, with its annotation, into an unlabeled volume, replacing the stagnant part of the boundary supervision with ground truth that is correct by construction \cite{bai2023bidirectional}. What AR and G cannot supply is geometry in the penalty itself: Homotopy-Annealed Hausdorff Supervision (HAH) anneals a distance-weighted loss from an annotation-anchored objective to a full Hausdorff surrogate \cite{karimi2019reducing}, admitting its prediction-dependent term only once the predicted surface is reliable, so that this term does not act while AR and G are still stabilizing the targets. HAH is the only component that modifies the objective; AR and G remain loss-free and leave the target unchanged when their trigger is absent.

Methods validated only on small benchmarks prove little about generality, and the semi-supervised pancreas literature lives on a 62-volume NIH training pool (12 labelled) \cite{roth2015deeporgan} and a 224-volume MSD pool \cite{antonelli2022medical}, where gains are fragile to split and seed. We therefore derive PanTS-SSL from the public PanTS corpus \cite{li2026pants}: a curated protocol with a homogeneous portal-venous main pool, labelled tiers of 50, 100 and 250 volumes, a tumor track stratified by lesion size, seed replicates, and probes for contrast phase, slice thickness and organ atrophy, scored against a fixed 901-volume test set touched once. The images and masks belong to PanTS; our contribution is the protocol, which measures how gains change with the label budget on a fixed, unscreened test set and releases shift probes, in the spirit of standardized benchmarks \cite{isensee2021nnu}.

In summary, the full method reaches $84.60\%$ Dice and $4.97$ voxels 95HD on NIH with 12 labelled volumes, the best among the compared methods, and $78.90\%$ and $4.19$ on MSD Task07 with 112; on PanTS-SSL it exceeds the co-training baseline at 100 labels and reaches $85.29\%$ Dice at 250 labels with AR+G and $85.50\%$ at 100 labels with HAH. We contribute an anatomy-aware account of what limits confidence-gated co-training on pancreas CT, confirmed by audits of the gate and of the rejected fragments; AR, a confidence-independent, loss-free edit that relabels implausible fragments as confident background; G, a loss-free transplant of labelled surface content and its annotation; HAH, a distance-weighted loss annealed to a Hausdorff surrogate that admits its prediction-dependent term only once the prediction is reliable; and PanTS-SSL, a large-scale benchmark with labelled tiers, a tumor track, seed replicates and shift probes over a fixed test set, on which accuracy rises monotonically with the label budget.

\section{Related Work}
\label{sec:related}

\noindent\textbf{Consistency and co-training in 3D medical imaging.}
Enforcing agreement between perturbed views underpins modern semi-supervised segmentation, from temporal ensembling \cite{laine2016temporal} to the mean teacher \cite{tarvainen2017mean}, enriched with shape priors \cite{li2020shape} and dual-task consistency \cite{luo2021semi}; co-training maintains two mutually supervising networks \cite{qiao2018deep} to avoid collapsing a single teacher-student pair \cite{berthelot2019mixmatch}, and has been extended with mutual consistency \cite{wu2022mutual}, causal intervention \cite{miao2023caussl} and mutual correction \cite{wang2023mcf}. We are orthogonal to this scaffolding: we keep the translation-consistent pipeline \cite{liu2024translation} as published and modify only the unlabeled targets, plus the penalty where those targets are exact.

\noindent\textbf{Confidence gating and pseudo-label selection.}
Fixed and adaptive confidence thresholds are the standard pseudo-label
selector \cite{sohn2020fixmatch}, refined by Monte Carlo uncertainty
\cite{yu2019uncertainty} or calibration \cite{rizve2021defense}. All derive
the acceptance score from the prediction judged, which works when errors carry
low confidence; our audits show the dominant error here does not, so no
prediction-derived score separates it, motivating a criterion outside the
prediction. Relabeling implausible fragments as confident background aligns
with negative learning \cite{kim2019nlnl, ishida2017learning}, but we apply it dynamically
to the network's own topological errors during training rather than to a
fixed external label set.

\noindent\textbf{Region mixing and copy-paste augmentation.}
Region-level mixing has a long history: CutMix perturbations strengthen
consistency \cite{yun2019cutmix}, bidirectional copy-paste transfers content
between labeled and unlabeled distributions \cite{bai2023bidirectional}, and
later work explored adaptive displacement \cite{chi2024adaptive} and
multi-network mutual learning \cite{chen2023magicnet}. Our grafting uses the same mechanics to replace a targeted fraction of incorrect boundary supervision with transplanted ground truth, quantifying the change in supervision quality. Unlike BCP \cite{bai2023bidirectional}, which pastes in both directions inside a mean teacher and blends annotation with pseudo-labels, G pastes only labelled content, only into the unlabeled branch of co-training, as an exact target that the confidence gate always admits.

\noindent\textbf{Boundary-aware and distance-weighted objectives.}
Dice and cross-entropy are permutation-invariant in the voxel index, so a boundary displaced by a few voxels costs no more than an interior flip, which motivates distance-aware objectives.
Boundary loss reweights the error by the distance transform of the annotation
\cite{kervadec2021boundary}, and region-based losses such as the Tversky and
combo losses rebalance foreground against background
\cite{salehi2017tversky, taghanaki2019combo}, yet none scales the penalty by
the distance to the predicted surface. The Hausdorff surrogate of Karimi and Salcudean does so, scaling each voxel's error by its distance to both the ground-truth and the predicted surface \cite{karimi2019reducing}, but the prediction-dependent term is treated as constant during differentiation, so the update is not the gradient of any fixed functional and is dominated by an uninformative term when the prediction is unreliable. Fixed-weight Hausdorff losses thus trade stability for geometry and need per-dataset tuning \cite{ma2021loss}. HAH instead treats the surrogate as the endpoint of a homotopy from an annotation-anchored objective that is convex in the prediction, admitting the stop-gradient predicted-surface term only once the prediction is reliable and only on voxels with exact annotation, so it coexists with the pseudo-label edits that stabilize the early schedule.

\noindent\textbf{Anatomical priors and evaluation protocols.}
Anatomical knowledge usually enters through the objective, e.g.\ regressing a signed distance map \cite{li2020shape}; such priors are soft, compete with the data term, and need a per-dataset weight. Ours is a hard, parameter-free edit of the pseudo-label distribution, so the prior enters as supervision and, unlike an output-time cleanup, changes what the network learns. Connected component filtering is a standard inference-time step, also evaluated for our starting point \cite{liu2024translation}; AR applies a tolerant form of the same prior to training targets and relabels rather than discards (compared in Sec.~\ref{sec:ablations}). Finally, the field's empirical basis is narrow, relying on NIH \cite{roth2015deeporgan} and MSD Task07 \cite{antonelli2022medical}; standardized protocols drove progress in fully supervised segmentation \cite{isensee2021nnu}, and PanTS-SSL provides the same service for the semi-supervised setting \cite{li2026pants}.

\section{Method}
\label{sec:method}
\begin{figure*}[t]
\centering
\includegraphics[width=\linewidth]{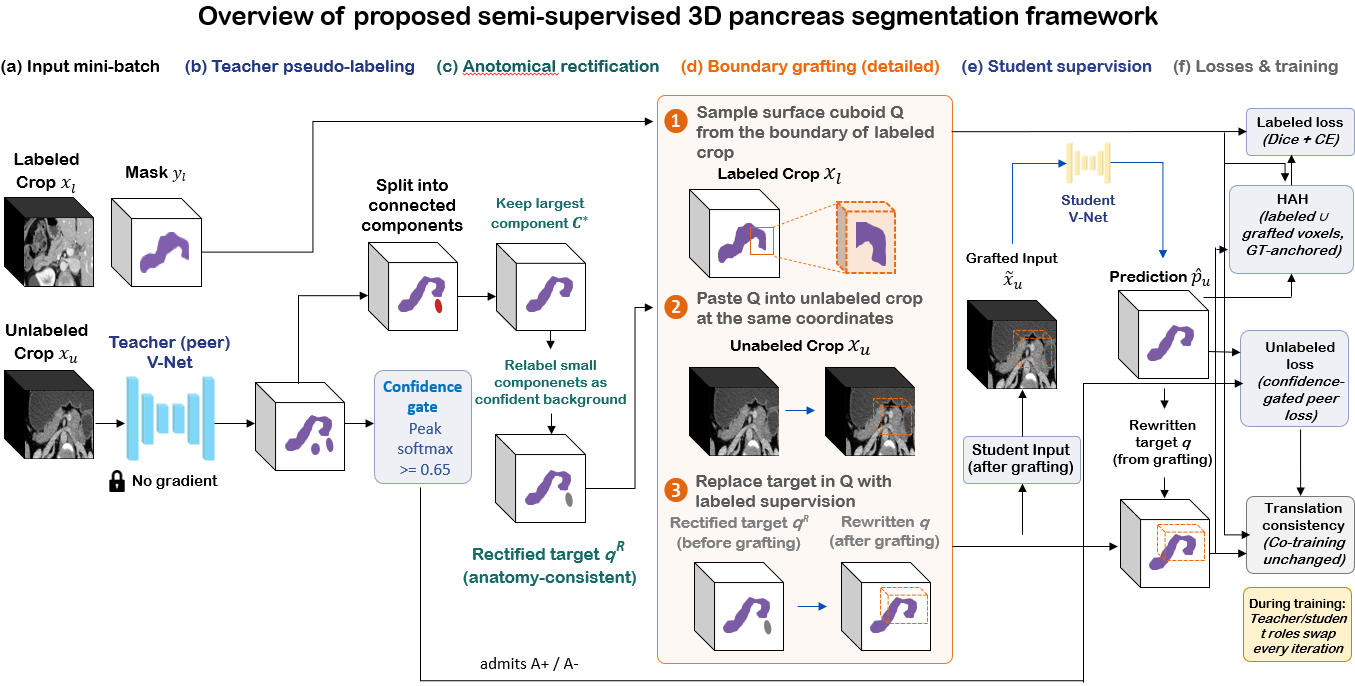}
\caption{\textbf{Overview.} The three components sit between the teacher
forward pass and the losses; the inherited objective is otherwise untouched.
\emph{Rectification} relabels fragments smaller than half the main component
as confident background. \emph{Grafting} transplants a cuboid $\mathcal{Q}$ of
a labeled crop, with its annotation, into the unlabeled input, so part of the
boundary supervision is ground truth by construction. \emph{HAH} anneals a
distance-weighted penalty on voxels carrying exact annotation.}
\label{fig:main}
\end{figure*}

Figure~\ref{fig:main} summarizes the design: three components act between the teacher forward pass and the losses of a confidence-gated co-training pipeline, leaving the architecture and inherited terms unchanged. Two V-Nets \cite{milletari2016v} $f_{\theta_1},f_{\theta_2}$ train with independent optimizers on $96^3$ crops, each batch holding $n_l=2$ labeled then $n_u=2$ unlabeled crops. For network $k$ with peer $j\neq k$, $\mathbf{z}^k$ is the logit map on a noisy input, $\sigma(\cdot)$ the channel softmax, $\omega$ a voxel (channel $1$ foreground). Labeled crops receive cross-entropy plus Dice; all crops receive a KL translation-consistency term, by which two overlapping crops at different offsets must agree on their intersection, with an entropy-balancing regularizer; unlabeled crops are supervised by the peer through the confident regional cross-entropy (CRC) of the starting point \cite{liu2024translation}, with admission gates
\begin{equation}
\begin{split}
\mathcal{A}^{+} &= \bigl\{\omega:\max_c \sigma(\mathbf{z}^j_\omega)_c \ge \gamma\bigr\}, \\
\mathcal{A}^{-} &= \bigl\{\omega:\min_c \sigma(\mathbf{z}^j_\omega)_c < \beta\bigr\},
\end{split}
\label{eq:gates}
\end{equation}
$\gamma=0.65$, $\beta=0.1$: voxels in $\mathcal{A}^{+}$ are positive targets weighted by peer confidence, those in $\mathcal{A}^{-}$ negative targets, the rest ignored. Network $2$ also receives 3D CutMix \cite{yun2019cutmix}, and the unsupervised weight follows a sigmoid ramp-up $\lambda_u(t)$. We add two loss-free, target-level mechanisms, Anatomical Rectification and Boundary Grafting (Secs.~\ref{sec:ar} and~\ref{sec:graft}), which replace the CRC target map ($\mathbf{z}^j$ in the starting point) on selected voxels with explicit confident log-probabilities, and one loss-level mechanism, Homotopy-Annealed Hausdorff Supervision (Sec.~\ref{sec:hah}), which reshapes how a correct target is penalized. They are motivated by the starting point on NIH, pooled over training: the gate admits $86$ to $91\%$ of the boundary-shell voxels of the annotation, yet only about $80\%$ of these admitted targets are correct.

\subsection{Anatomical Rectification}
\label{sec:ar}
The pancreas is a single connected structure, so a peer prediction that splits
into disjoint pieces is anatomically inconsistent however confident. Let
$M=\{\omega:\operatorname{argmax}_c\,\mathbf{z}^j_\omega=1\}$ on an unlabeled
crop, with 26-connected components $C_1,\dots,C_N$ and largest $C_\star$. For
tolerance $\rho\in(0,1]$, Rectification rejects
\begin{equation}
\mathcal{R}=\bigcup_{i\neq\star,\ |C_i|<\rho\,|C_\star|} C_i
\label{eq:reject}
\end{equation}
and writes a confident background target on those voxels,
\begin{equation}
\mathbf{q}^{\mathcal{R}}_\omega=
\begin{cases}
\bigl(\log\kappa,\ \log(1-\kappa)\bigr), & \omega\in\mathcal{R},\\[2pt]
\mathbf{z}^j_\omega, & \text{otherwise},
\end{cases}
\label{eq:rectify}
\end{equation}
$\kappa=0.99$, unlabeled half only. A rectified voxel lands in both $\mathcal{A}^{+}$ (background, weight $\kappa$) and $\mathcal{A}^{-}$, so both CRC branches push away from foreground: the network is told the organ is absent (negative learning \cite{ishida2017learning}), not told nothing. AR leaves the target unchanged when $N\le1$ or no component falls below the tolerance, reads topology rather than confidence, and uses $\rho=0.5$ as a tolerance, not a knob: the strict largest-component variant ($\rho=1$) deletes a legitimate piece whenever a $96^3$ crop clips the elongated pancreas into two comparable pieces, degrading both metrics. It is not free of risk: a true piece smaller than $\rho|C_\star|$ becomes background, which the audit (Sec.~\ref{sec:ablations}) finds for $11.2\%$ of rejected components, although it also shows the rule mostly removes hallucinations rather than anatomy; AR trades a little label noise for the removal of confident hallucinations.

\subsection{Boundary Grafting}
\label{sec:graft}

Rectification cannot repair the roughly one fifth of admitted shell targets that is incorrect (Sec.~\ref{sec:method}); this error is correlated across the two networks, so no reweighting recovers it. Grafting supplies the missing information by transplanting annotation: for unlabeled crop $u$, labeled crop $s$ and cuboid $\mathcal{Q}$,
\begin{align}
\tilde{\mathbf{x}}_u(\omega) &=
\begin{cases}
\mathbf{x}_s(\omega), & \omega\in\mathcal{Q},\\
\mathbf{x}_u(\omega), & \text{otherwise},
\end{cases}
\label{eq:graft-x}\\[2pt]
\mathbf{q}_\omega &=
\begin{cases}
\mathrm{onehot}_\kappa\bigl(\mathbf{y}_s(\omega)\bigr), & \omega\in\mathcal{Q},\\
\mathbf{q}^{\mathcal{R}}_\omega, & \text{otherwise},
\end{cases}
\label{eq:graft-t}
\end{align}
$\mathrm{onehot}_\kappa(c)$ places $\log\kappa$ on channel $c$. Grafted voxels carry annotation (no label noise) and the source is always labeled. The centre of $\mathcal{Q}$ is drawn uniformly on the surface $G_s\setminus\mathrm{ero}(G_s)$, extents are 48 to 72 voxels, one graft per crop (skipped if $G_s$ is empty); on network $2$ it follows CutMix, sourcing from the original labeled crops. Surface placement spends the budget where targets are worst: one graft covers about $23\%$ of a crop and converts about $61\%$ of its shell supervision to ground truth.

\begin{table*}[!t]
\centering
\caption{Results on the NIH dataset (12 labeled) and MSD Task07 (112 labeled).
All methods use the same splits (NIH: 12 labeled, 50 unlabeled, 20 test; MSD: 112 labeled, 112 unlabeled, 57 test); distances in voxels, --: not reported. Best results in \textbf{bold}.}
\label{tab:panc}
\resizebox{\textwidth}{!}{%
\begin{tabular}{lcccc|cccc}
\hline\hline
 & \multicolumn{4}{c|}{NIH} & \multicolumn{4}{c}{MSD Task07} \\
\cline{2-9}
Method
 & DSC(\%)$\uparrow$ & Jacc(\%)$\uparrow$ & 95HD$\downarrow$ & ASD$\downarrow$
 & DSC(\%)$\uparrow$ & Jacc(\%)$\uparrow$ & 95HD$\downarrow$ & ASD$\downarrow$ \\
\hline
MT \cite{tarvainen2017mean}
 & 74.43 & 60.53 & 14.93 & 4.61 & 73.42 & 58.96 & 15.93 & 4.78 \\
UA-MT \cite{yu2019uncertainty}
 & 76.10 & 62.62 & 10.84 & 2.43 & 73.58 & 59.08 & 15.62 & 4.63 \\
SASSNet \cite{li2020shape}
 & 76.39 & 63.17 & 11.06 & 1.42 & 73.09 & 58.51 & 14.87 & 4.29 \\
DTC \cite{luo2021semi}
 & 76.27 & 62.82 & 8.70 & 2.20 & 72.89 & 58.28 & 14.43 & 4.18 \\
MC-Net \cite{wu2021semi}
 & 78.17 & 65.22 & 6.90 & 1.55 & 73.70 & 59.24 & 13.45 & 4.02 \\
MCF \cite{wang2023mcf}
 & 79.21 & 66.09 & 6.56 & 1.76 & 73.82 & 59.38 & 13.12 & 3.90 \\
Contrastive \cite{karimijafarbigloo2024reducing}
 & 79.42 & 66.33 & 7.55 & 3.10 & 74.05 & 59.54 & 12.87 & 3.75 \\
MC-Net+ \cite{wu2022mutual}
 & 79.37 & 66.83 & 8.52 & 1.72 & 73.98 & 59.52 & 13.08 & 3.80 \\
MCCauSSL \cite{miao2023caussl}
 & 80.92 & 68.26 & 8.11 & 1.53 & 74.32 & 59.82 & 12.21 & 3.58 \\
PS-MT \cite{liu2022perturbed}
 & 80.74 & 68.15 & 7.41 & 2.06 & 74.18 & 59.68 & 12.43 & 3.69 \\
TraCoCo \cite{liu2024translation}
 & 81.80 & 69.56 & 5.70 & 1.49 & 75.75 & 61.93 & 6.58 & 2.09 \\
SDRML \cite{xie2026spatially}
 & 80.31 & 67.61 & 10.59 & 3.03 & -- & -- & -- & -- \\
DMC \cite{saleknia2026distributional}
 & 82.50 & 70.49 & 5.52 & 1.35 & 76.01 & 62.58 & 5.16 & 1.61 \\
UnCoL \cite{uncol2025}
 & 83.16 & 71.46 & 5.44 & 1.44 & -- & -- & -- & -- \\
\hline
\rowcolor[HTML]{C8FFFD}
\textbf{Ours}
 & \textbf{84.60} & \textbf{73.50} & \textbf{4.97} & \textbf{1.29}
 & {\textbf{78.90}} & {\textbf{65.85}} & {\textbf{4.19}} & {\textbf{1.19}} \\
\hline\hline
\end{tabular}}
\end{table*}

\subsection{Homotopy-Annealed Hausdorff Supervision}
\label{sec:hah}
Rectification and Grafting determine \emph{what} the student fits, not \emph{how} a correct target is penalized. Once AR and G remove coarse hallucinations, the residual error is coherent small-scale surface displacement (Fig.~\ref{fig:qual3}), which permutation-invariant Dice and cross-entropy penalize no more than an interior flip and a distance-weighted objective addresses directly. The Hausdorff surrogate of Karimi and Salcudean \cite{karimi2019reducing} scales each voxel's error by its distance to both surfaces, but treats its predicted-side term as constant in the gradient, which is then large and uninformative for unreliable predictions, precisely when AR and G stabilize the pseudo-label. HAH is therefore a refinement on top of AR and G, not a standalone loss, and is evaluated that way (Sec.~\ref{sec:ablations}).

We instead use a continuation schedule. With $\tau=t/T\in[0,1]$, $p_{\theta_k}(\omega)=\sigma(\mathbf{z}^k_\omega)_1$, $d_Y$ the ground-truth distance map of a labeled crop's annotation $g$ (computed once per crop) and $d_{P_{\theta_k}}$ the distance map of the current hard segmentation, both normalized by their per-crop maximum, we define
\begin{equation}
\begin{split}
\mathcal{L}_{\mathrm{HAH}}(\theta_k;\tau) =
\frac{1}{|\Omega_g|}\sum_{\omega\in\Omega_g}
\big(p_{\theta_k}(\omega)-g(\omega)\big)^2 \\
\times \Big[d_Y(\omega)^\alpha +
w(\tau)\cdot \mathrm{sg}\big(d_{P_{\theta_k}}(\omega)^\alpha\big)\Big],
\end{split}
\label{eq:hah}
\end{equation}
with stop-gradient $\mathrm{sg}[\cdot]$ on the predicted-side term, schedule $w(\tau)=\tau$, exponent $\alpha=2$, and $\Omega_g$ the voxels carrying exact ground truth (the labeled crops that contain foreground), where $d_Y$ is well defined. HAH is added to the supervised loss, $\mathcal{L}_{\mathrm{sup}}=\tfrac12(\ell_{\mathrm{ce}}+\mathcal{L}_{\mathrm{Dice}})+\lambda\,\mathcal{L}_{\mathrm{HAH}}$ with $\lambda=0.3$; CRC and translation consistency are unaffected. At $\tau=0$,
\begin{equation}
\mathcal{L}_{\mathrm{HAH}}(\theta_k;0)=
\frac{1}{|\Omega_g|}\sum_{\omega\in\Omega_g}
\big(p_{\theta_k}(\omega)-g(\omega)\big)^2 d_Y(\omega)^\alpha,
\label{eq:hah0}
\end{equation}
a fixed functional depending only on the annotation, convex in the prediction $p$ though not in $\theta_k$, with unique minimizer $p^\star=g$ over $p$ and an exact gradient. At $\tau=1$ the full surrogate is recovered; since $d_Y$ does not depend on $\tau$, both endpoints share $p^\star=g$, so the schedule reshapes the trajectory without moving its destination. Formally, $\mathcal{L}_{\mathrm{HAH}}(\cdot;\tau)$ is the straight-line homotopy between these two objectives \cite{allgower1990numerical}; in implementation it is a linear ramp of one term. Early on, $w(\tau)\approx0$ suppresses the high-variance predicted-surface term while AR and G stabilize the targets; as $\tau\to1$ it returns once the prediction is reliable, refining the boundary with a distance weighting that Dice and cross-entropy lack. HAH is the only component acting on the objective, uses the same constants ($w(\tau)=\tau$, $\alpha=2$, $\lambda=0.3$) on every dataset, and acts only where $d_Y$ is exact, so it inherits none of the label noise AR and G circumvent.

\paragraph{The full objective.} The objective of network $k$ is the starting-point objective with $(\mathbf{x}_u,\mathbf{z}^j)$ replaced by $(\tilde{\mathbf{x}}_u,\mathbf{q})=\mathcal{G}(\mathcal{R}(\mathbf{z}^j))$ inside the CRC term, for Rectification $\mathcal{R}$ and Grafting $\mathcal{G}$, plus $\lambda\mathcal{L}_{\mathrm{HAH}}$, with no learnable parameter; if the peer prediction is connected and no labeled foreground is present it reduces exactly to the baseline, otherwise AR can relabel a true organ piece (Sec.~\ref{sec:ar}).

\section{Experiments and Results}
\label{sec:experiments}

\begin{figure*}[!t]
\centering\small
\setlength{\tabcolsep}{2pt}
\begin{tabular}{@{}ccccc@{}}
\includegraphics[width=0.19\linewidth,height=2.6cm,keepaspectratio]{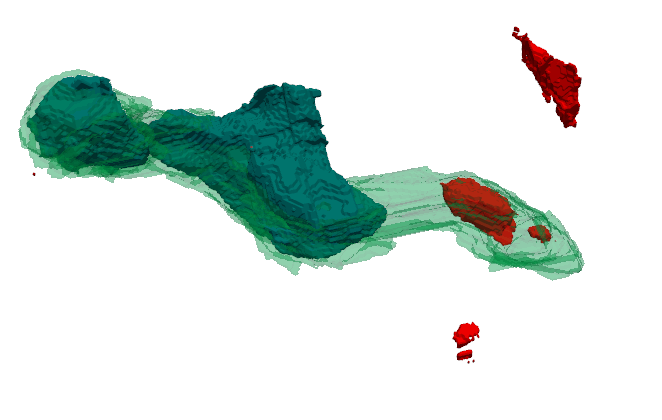} &
\includegraphics[width=0.19\linewidth,height=2.6cm,keepaspectratio]{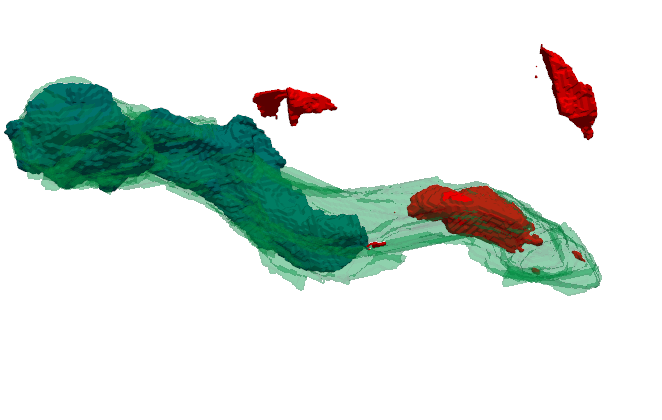} &
\includegraphics[width=0.19\linewidth,height=2.6cm,keepaspectratio]{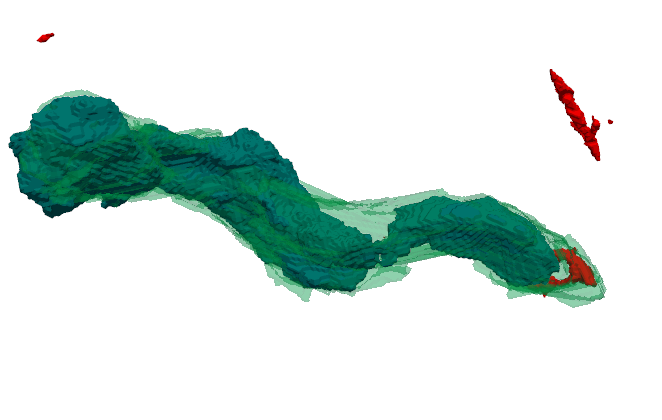} &
\includegraphics[width=0.19\linewidth,height=2.6cm,keepaspectratio]{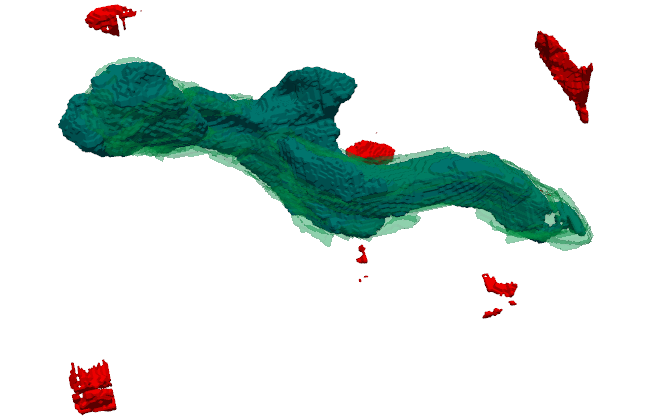} &
\includegraphics[width=0.19\linewidth,height=2.6cm,keepaspectratio]{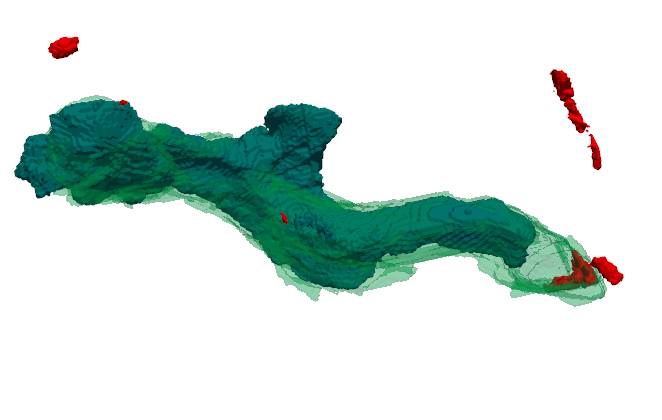} \\
{\scriptsize 1{,}000} & {\scriptsize 2{,}000} & {\scriptsize 3{,}000} & {\scriptsize 4{,}000} & {\scriptsize 5{,}000} \\[4pt]
\includegraphics[width=0.19\linewidth,height=2.6cm,keepaspectratio]{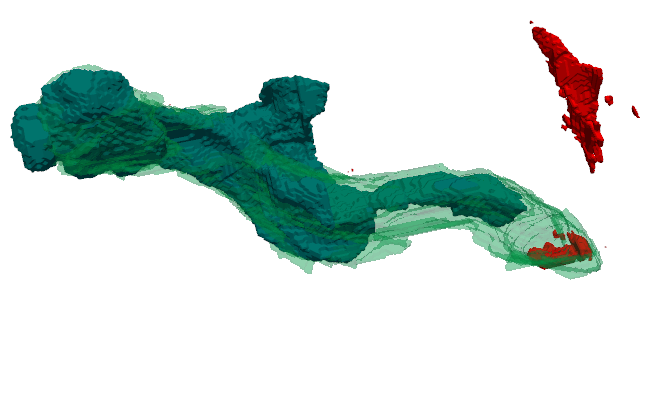} &
\includegraphics[width=0.19\linewidth,height=2.6cm,keepaspectratio]{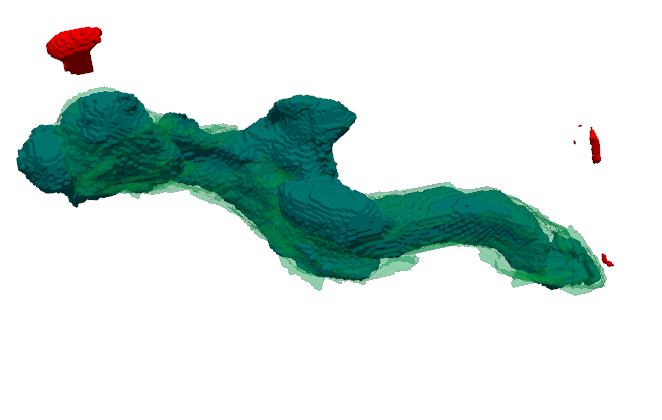} &
\includegraphics[width=0.19\linewidth,height=2.6cm,keepaspectratio]{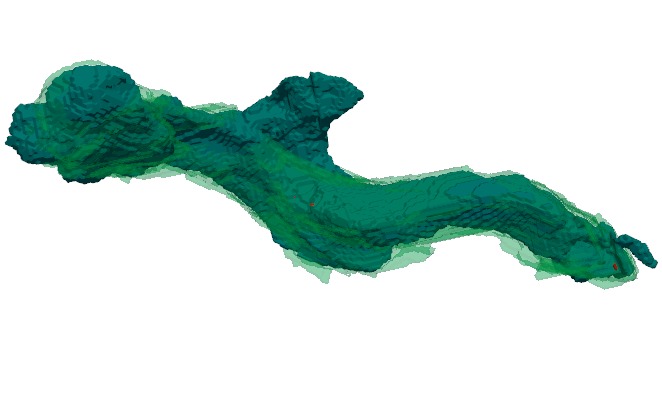} &
\includegraphics[width=0.19\linewidth,height=2.6cm,keepaspectratio]{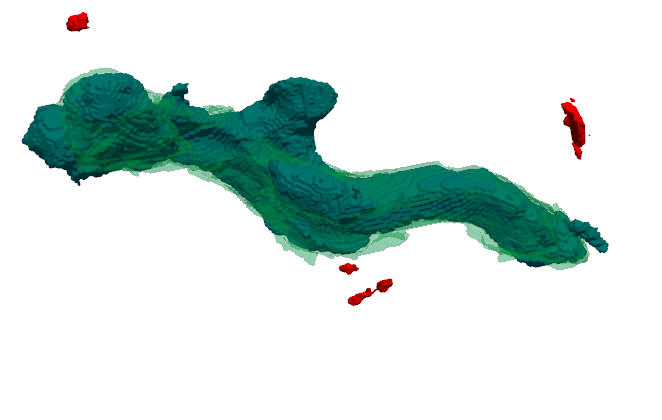} &
\includegraphics[width=0.19\linewidth,height=2.6cm,keepaspectratio]{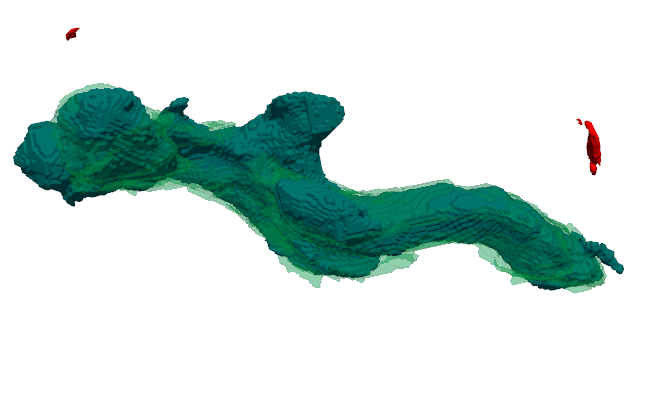} \\
{\scriptsize 6{,}000} & {\scriptsize 7{,}000} & {\scriptsize 8{,}000} & {\scriptsize 9{,}000} & {\scriptsize 10{,}000} \\
\end{tabular}
\caption{3D evolution of Ours on a held-out NIH case; detached fragments in red. Fragments do not vanish monotonically: they fluctuate (for example, around 6{,}000 iterations) but shrink over training, leaving a single small fragment at the final iteration; iteration index below each column. }
\label{fig:evol}
\end{figure*}

\subsection{Setup}
\label{sec:setup}
\noindent\textbf{Datasets.} \textbf{NIH Pancreas} \cite{roth2015deeporgan} has $82$ volumes; following \cite{liu2024translation} (their fold 0), we train on $62$ ($12$ labeled, $50$ unlabeled) and test on the held-out $20$, calling it the NIH dataset. \textbf{MSD Task07} \cite{antonelli2022medical} supplies $224$ training volumes ($112$ labeled, $112$ unlabeled) and $57$ for testing. On both, CT intensities are clipped to $[-120, 240]$ HU and normalized, and on MSD pancreas and tumor labels are merged into one pancreas class, so all methods in Table~\ref{tab:panc} share splits and a whole-organ label definition. \textbf{PanTS-SSL}, our large-scale benchmark, is introduced below.

\noindent\textbf{Implementation.} We keep every setting of the starting point \cite{liu2024translation}: two V-Nets \cite{milletari2016v} (16 base filters doubled at each of four down-sampling stages, batch normalization, dropout, 9.45 M parameters, as in \cite{yu2019uncertainty,liu2024translation}), independent SGD with learning rate $2.5\times10^{-2}$, momentum $0.9$, weight decay $10^{-4}$, $10{,}000$ iterations on $96^3$ crops, batches of four split evenly, $\gamma=0.65$, $\beta=0.1$, a $40$-epoch sigmoid ramp-up, and translation-consistency and entropy weights $1.0$ and $0.1$. Our constants are fixed on every dataset: $\rho=0.5$, $\kappa=0.99$, $26$-connectivity, graft extent 48 to 72 voxels, one graft per unlabeled crop, $w(\tau)=\tau$, $\alpha=2$, $\lambda=0.3$. Seed $1337$, one GPU per run; code, PanTS-SSL splits and the selection script will be released.

\noindent\textbf{Evaluation.} Inference is sliding-window with $96^3$ patches at stride $16$; we report Dice (DSC), Jaccard, 95th-percentile Hausdorff distance (95HD) and average surface distance (ASD), with distances in voxels as in \cite{liu2024translation}. As there, each configuration reports the last-iteration checkpoint; of the two co-trained networks we report the better by DSC, and since NIH and MSD have no validation split, the weaker network of the full method ($83.90$/$72.60$/$5.18$/$1.34$ on NIH, $78.35$/$65.15$/$4.75$/$1.30$ on MSD) is reported too and still beats every compared method in Table~\ref{tab:panc} on all four metrics. PanTS-SSL selects on its own validation set and is scored once on its test set. No post-processing (e.g., connected-component filtering) is applied, as in \cite{liu2024translation,saleknia2026distributional}. Table~\ref{tab:graft} is a separate sweep with Rectification disabled, compared only within itself; its $23\%$ surface row is the G-only configuration of Table~\ref{tab:abl} trained again, so they differ by run-to-run variation.

\noindent\textbf{PanTS-SSL (contribution).} The public PanTS corpus \cite{li2026pants} has $9{,}901$ CT studies annotated for organ and lesions but no semi-supervised protocol. A homogeneous main pool of $1{,}015$ volumes is selected by acquisition criteria (portal-venous phase, slice thickness $\le3$ mm, in-plane spacing $\le1$ mm, pancreas volume 35 to 150 cm$^3$, an intensity sanity window, an essentially single-component annotation, i.e.\ the second-largest annotated component holds at most $5\%$ of the organ volume, patient-level de-duplication). From it we draw a $100$-volume validation set, nested labeled tiers of 50, 100 and 250, an $80$-case tumor track stratified by lesion size, seed replicates, and evaluation-only shift probes (non-contrast, arterial, thick-slice, atrophic; $4{,}161$, $400$, $300$ and $200$ volumes), all scored against a fixed $901$-volume test set touched once. The component criterion filters the training pool only, removing annotations with large detached pieces that would make the labelled supervision inconsistent; because it agrees with the prior of AR, the test set (the official PanTS test split, taken without selection) and the probes are not connectivity-filtered, so evaluation does not select cases that suit the prior. The probes are released for robustness studies; no probe result is reported or enters any design decision.

\begin{figure*}[!t]
\centering\small
\setlength{\tabcolsep}{1pt}
\begin{tabular}{@{}ccccccc@{}}
\includegraphics[width=2.35cm,height=2.2cm]{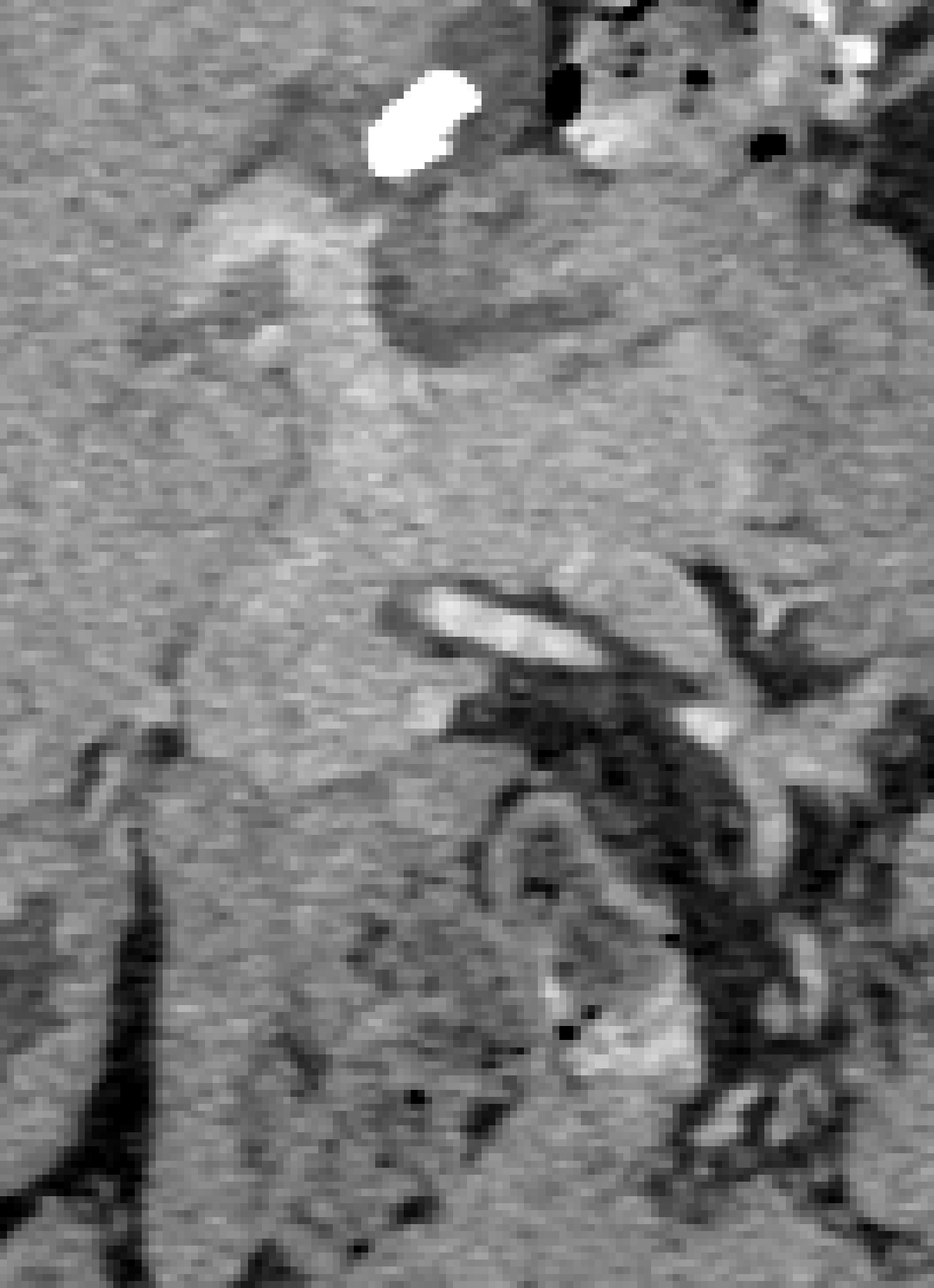} &
\includegraphics[width=2.35cm,height=2.2cm]{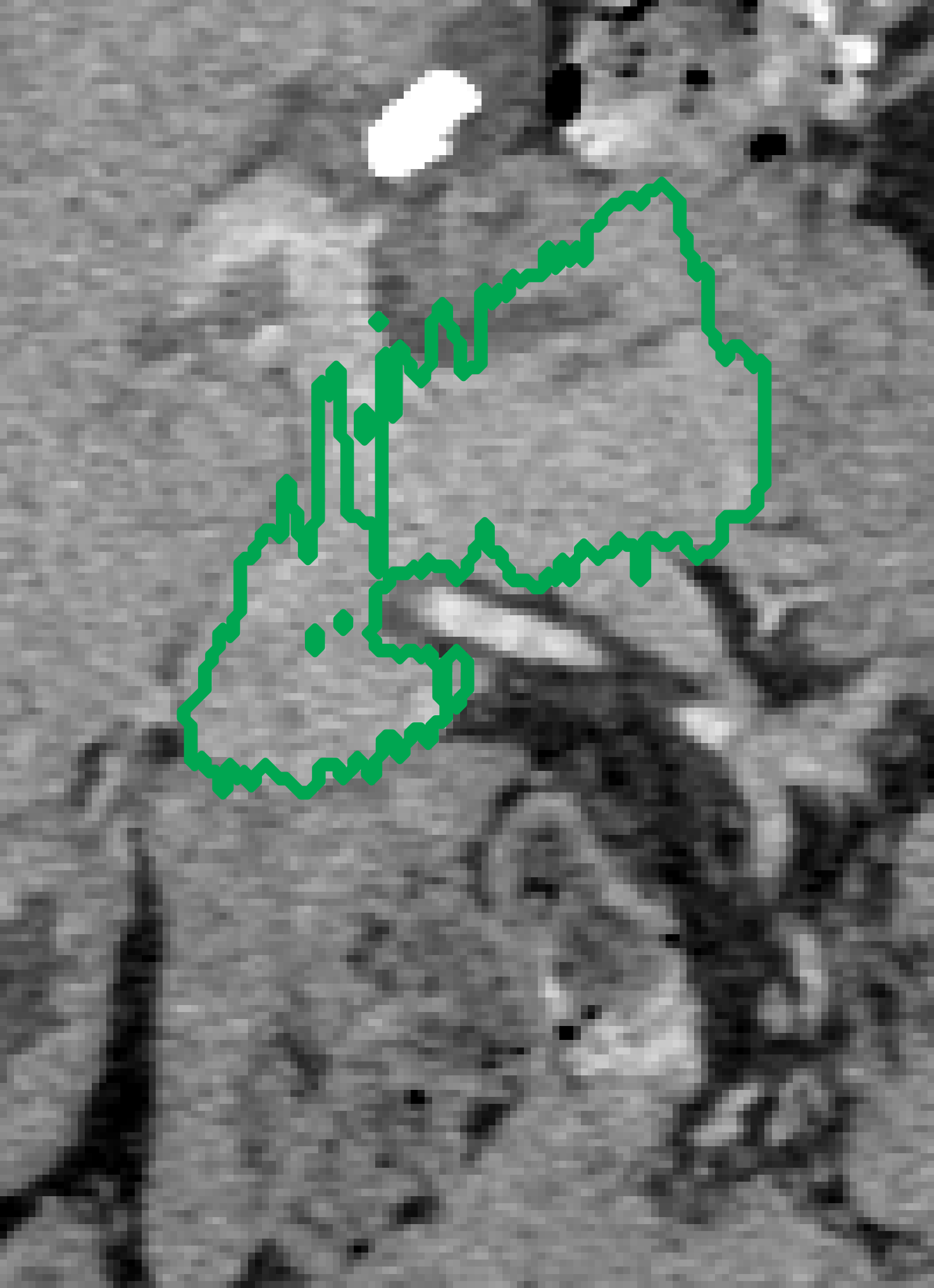} &
\includegraphics[width=2.35cm,height=2.2cm]{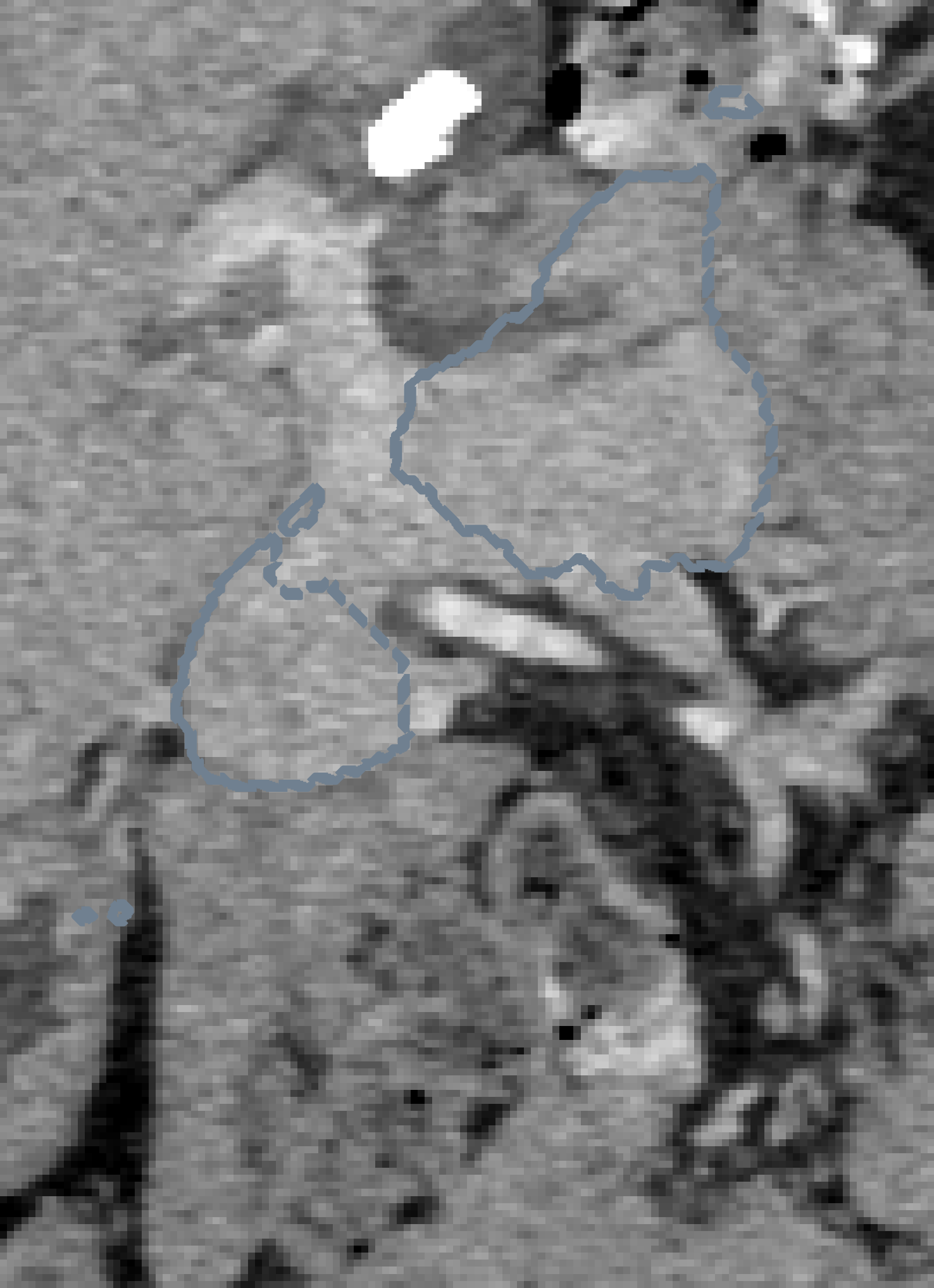} &
\includegraphics[width=2.35cm,height=2.2cm]{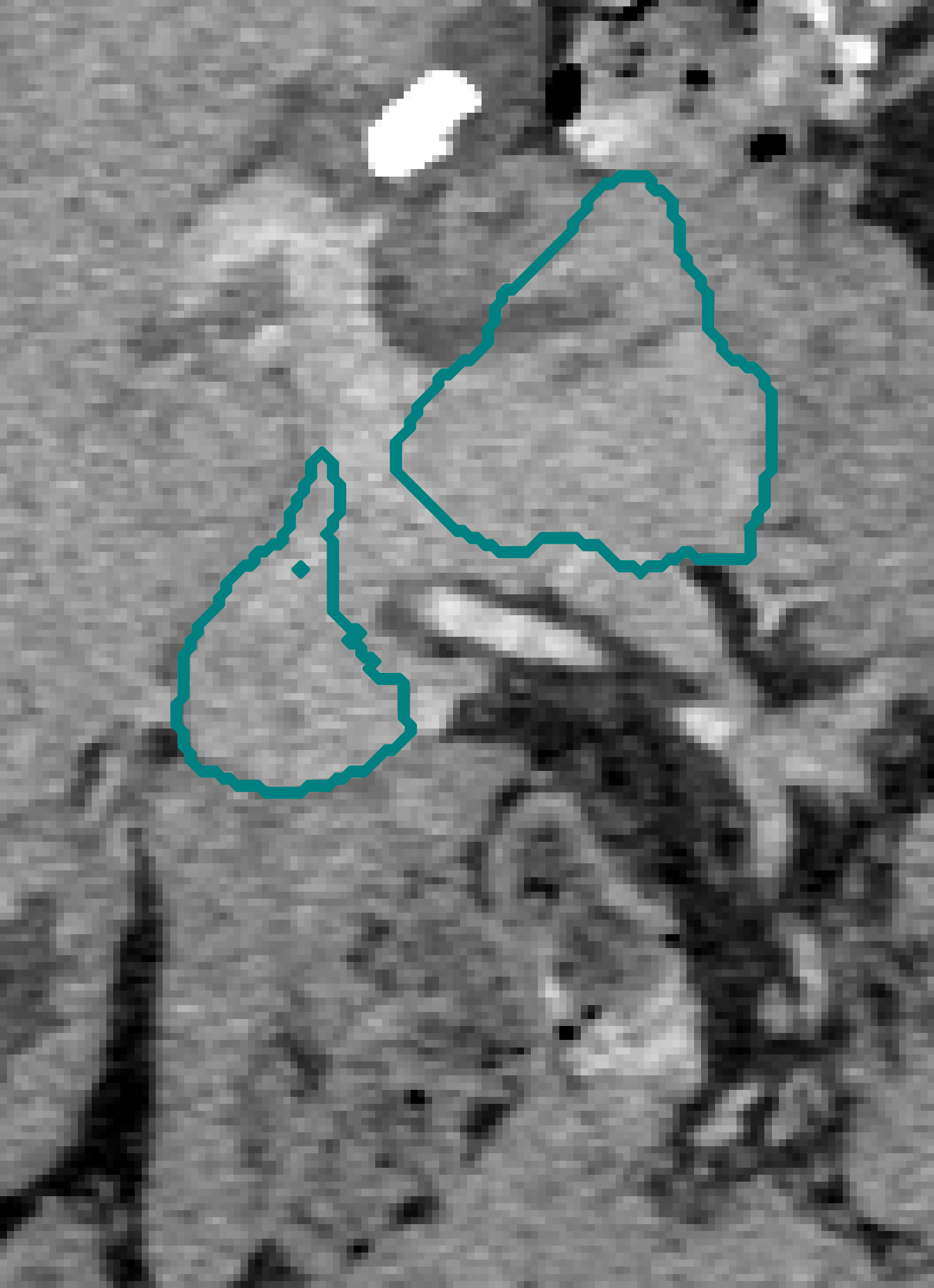} &
\includegraphics[width=2.35cm,height=2.2cm]{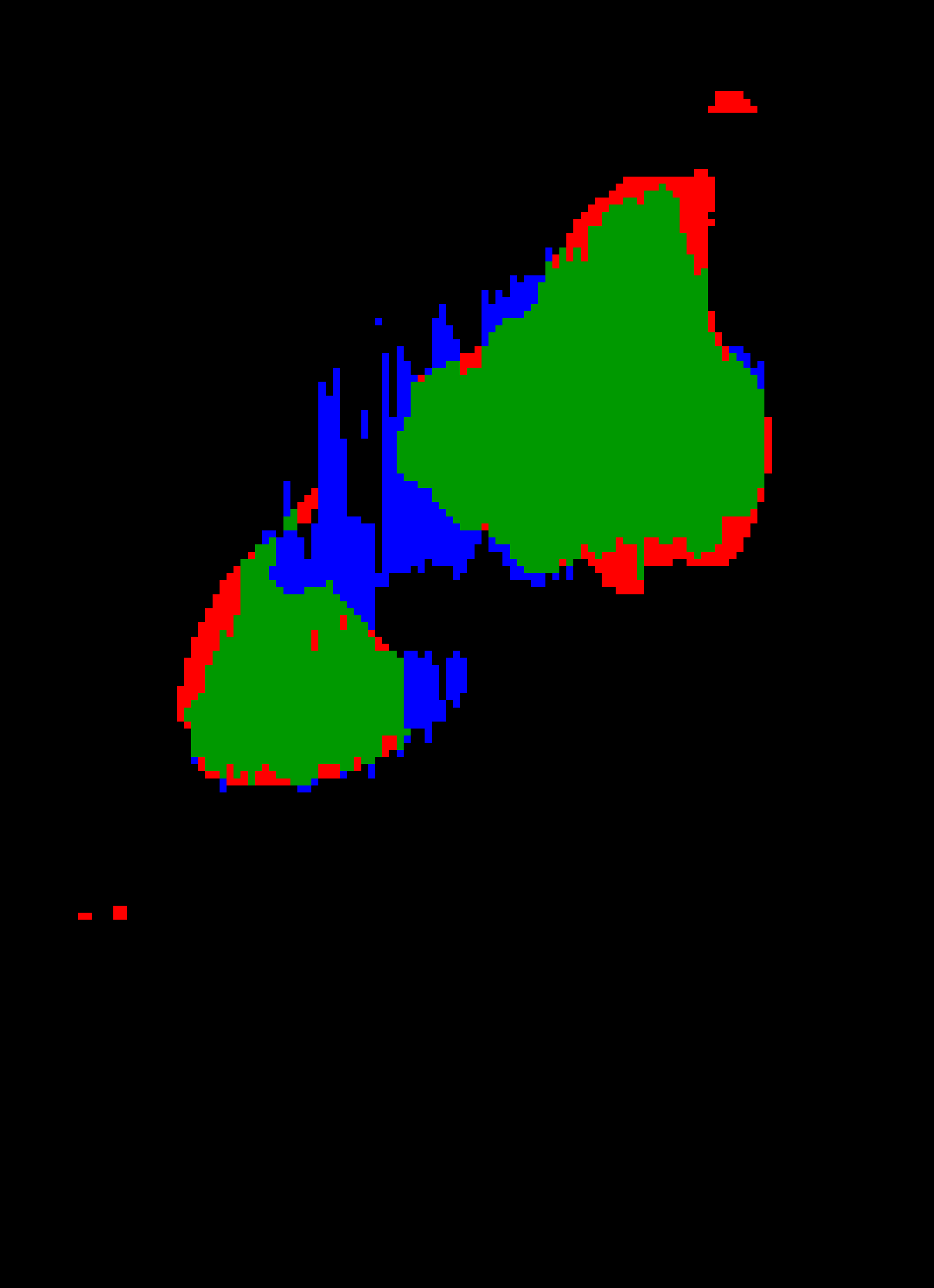} &
\includegraphics[width=2.35cm,height=2.2cm]{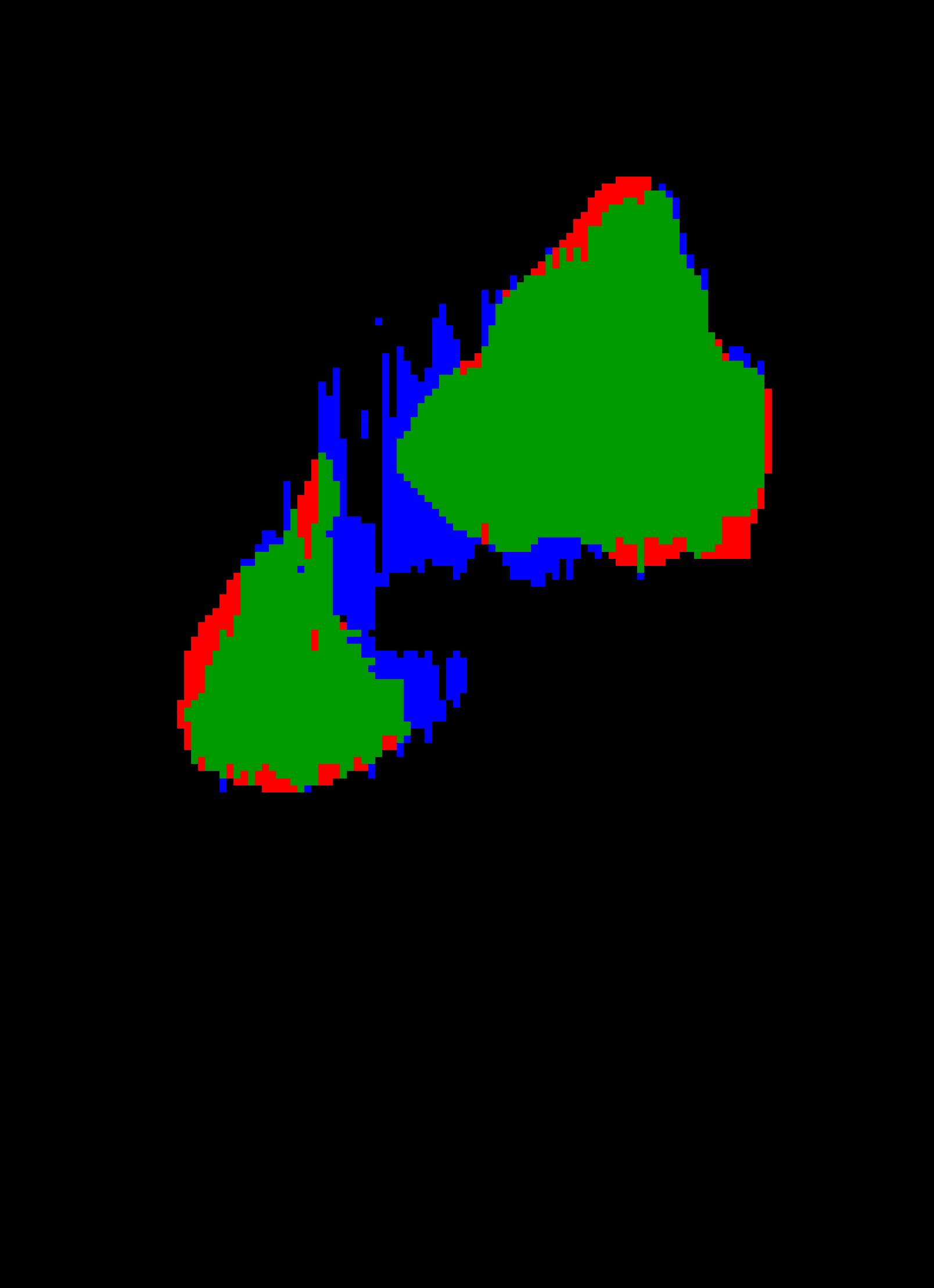} &
\includegraphics[width=2.35cm,height=2.2cm]{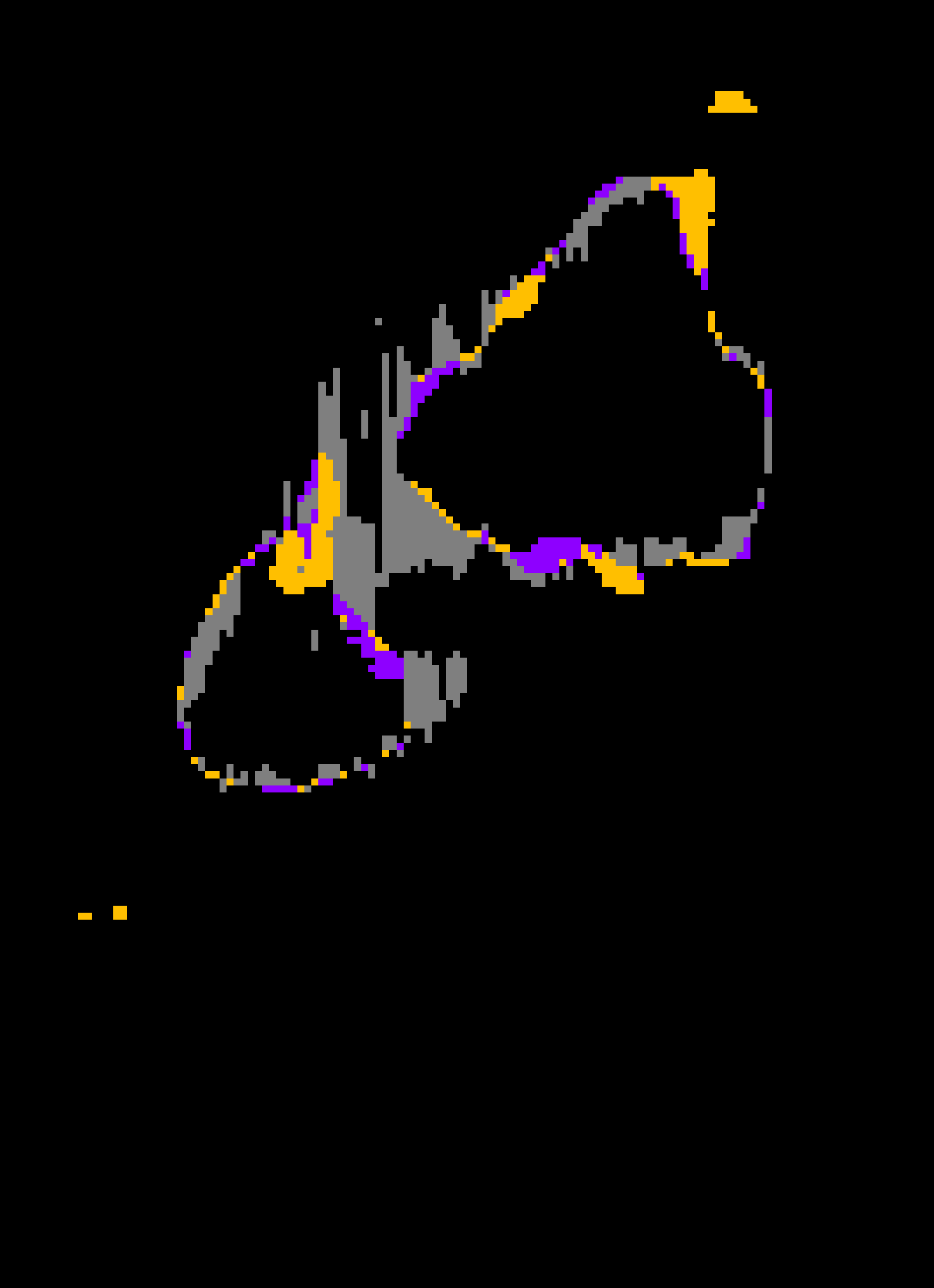} \\
\includegraphics[width=2.35cm,height=2.2cm]{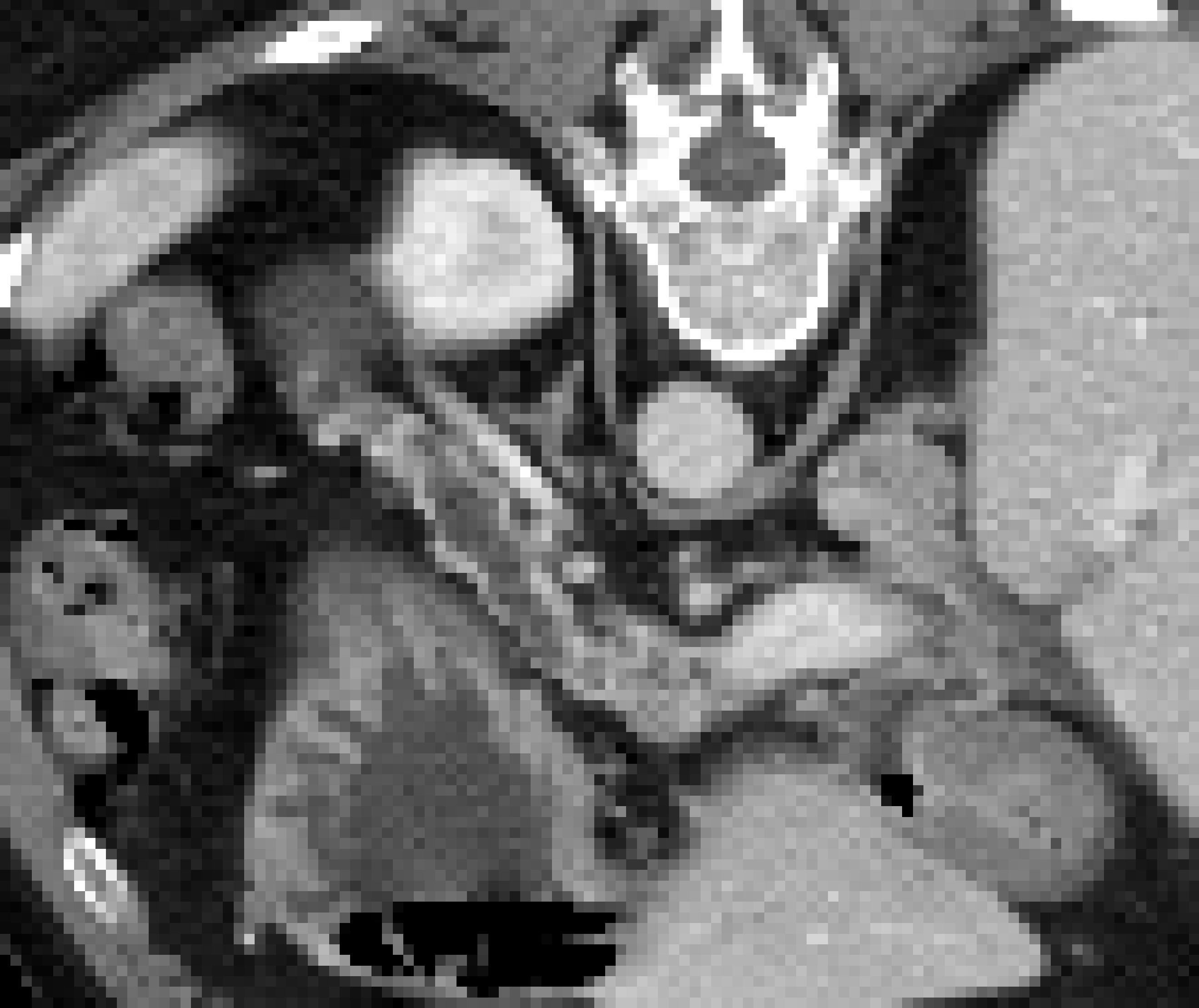} &
\includegraphics[width=2.35cm,height=2.2cm]{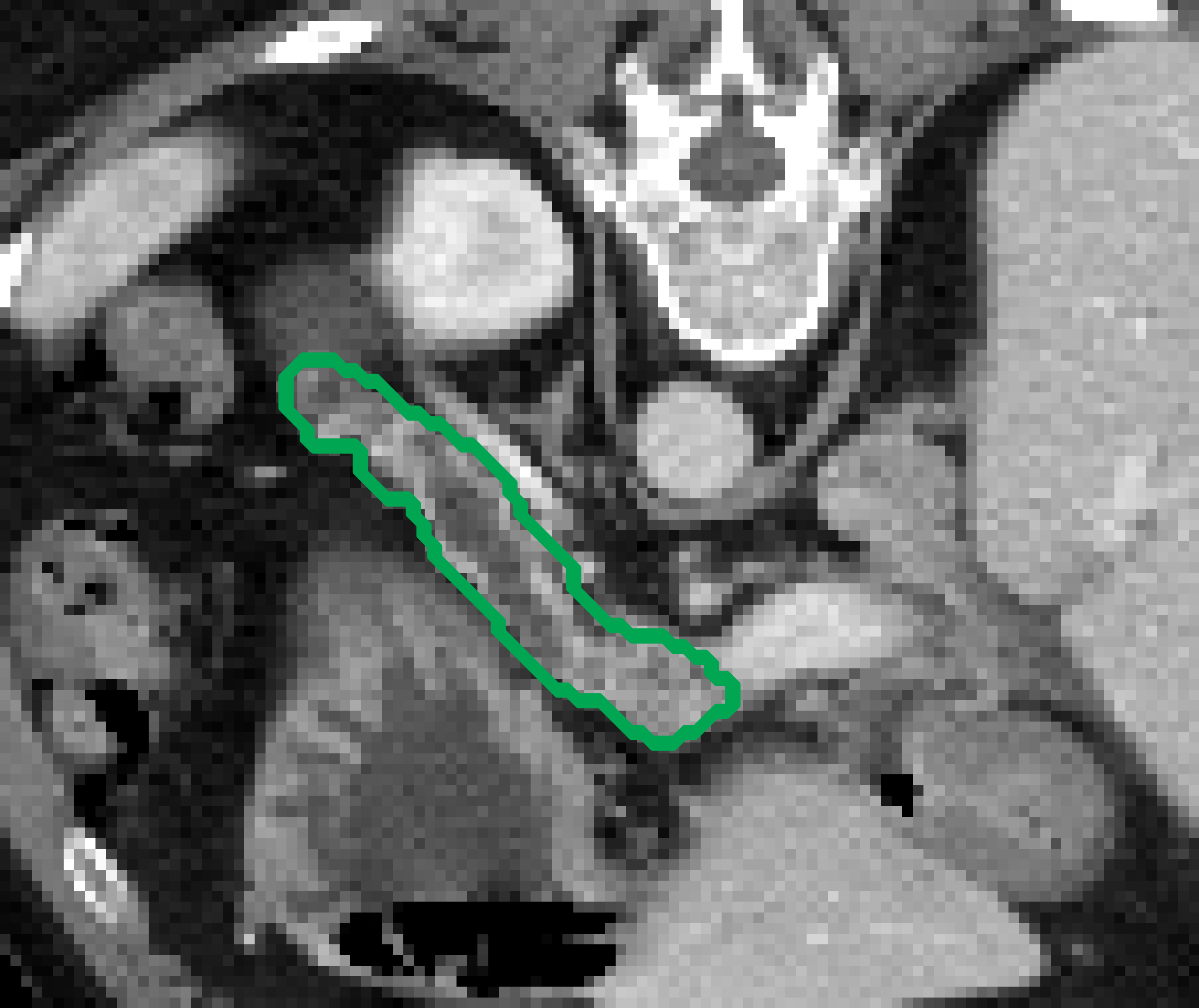} &
\includegraphics[width=2.35cm,height=2.2cm]{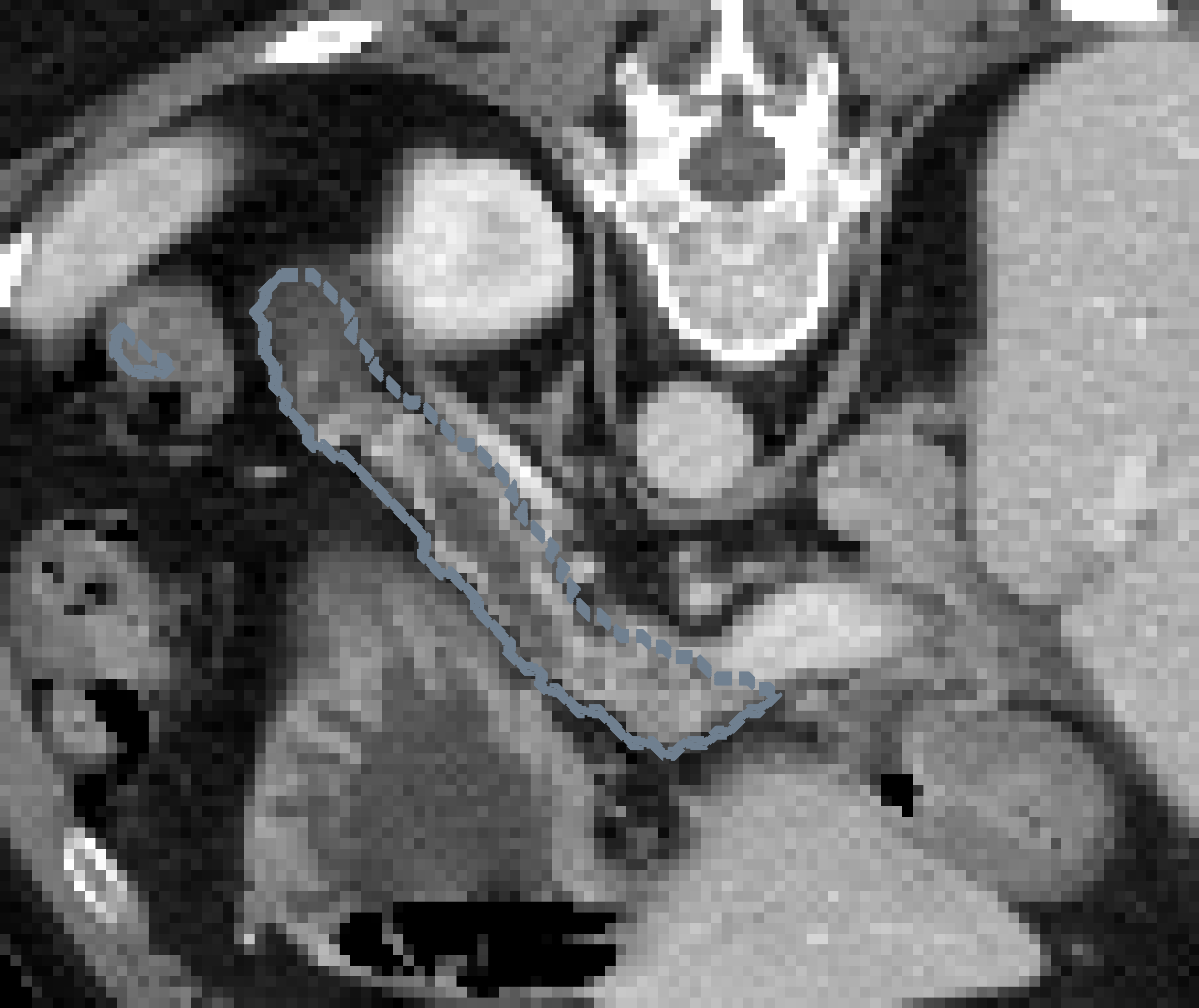} &
\includegraphics[width=2.35cm,height=2.2cm]{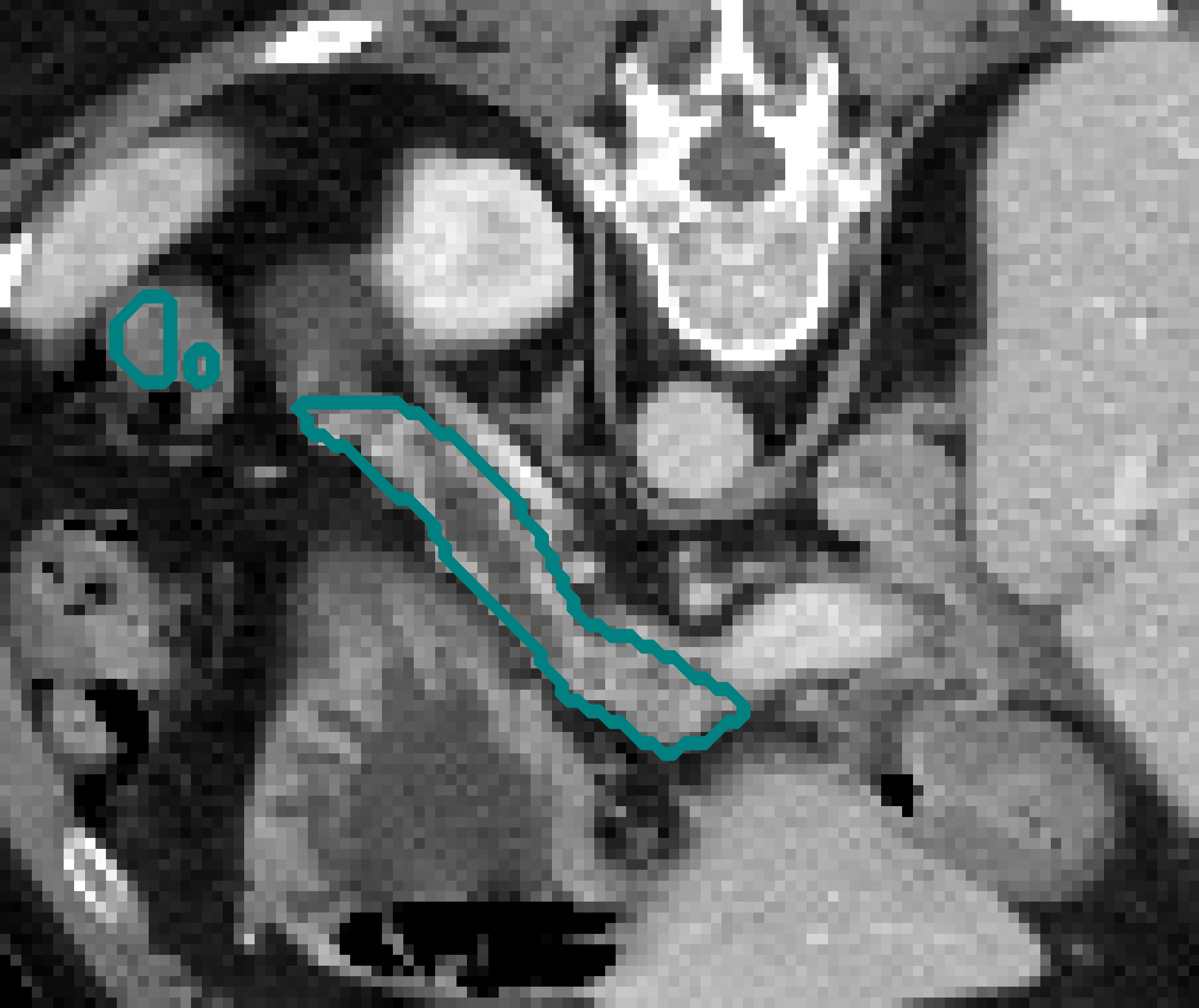} &
\includegraphics[width=2.35cm,height=2.2cm]{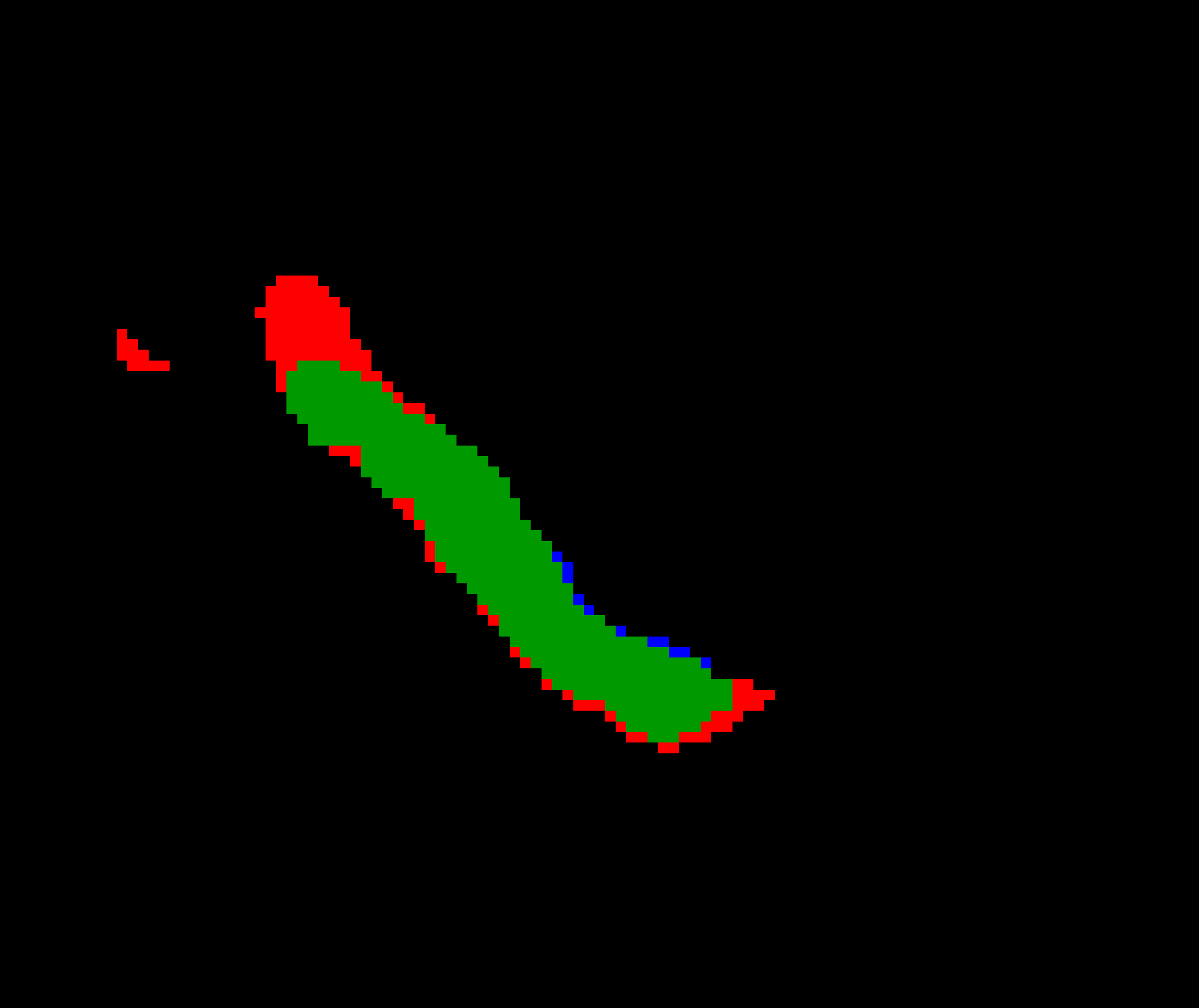} &
\includegraphics[width=2.35cm,height=2.2cm]{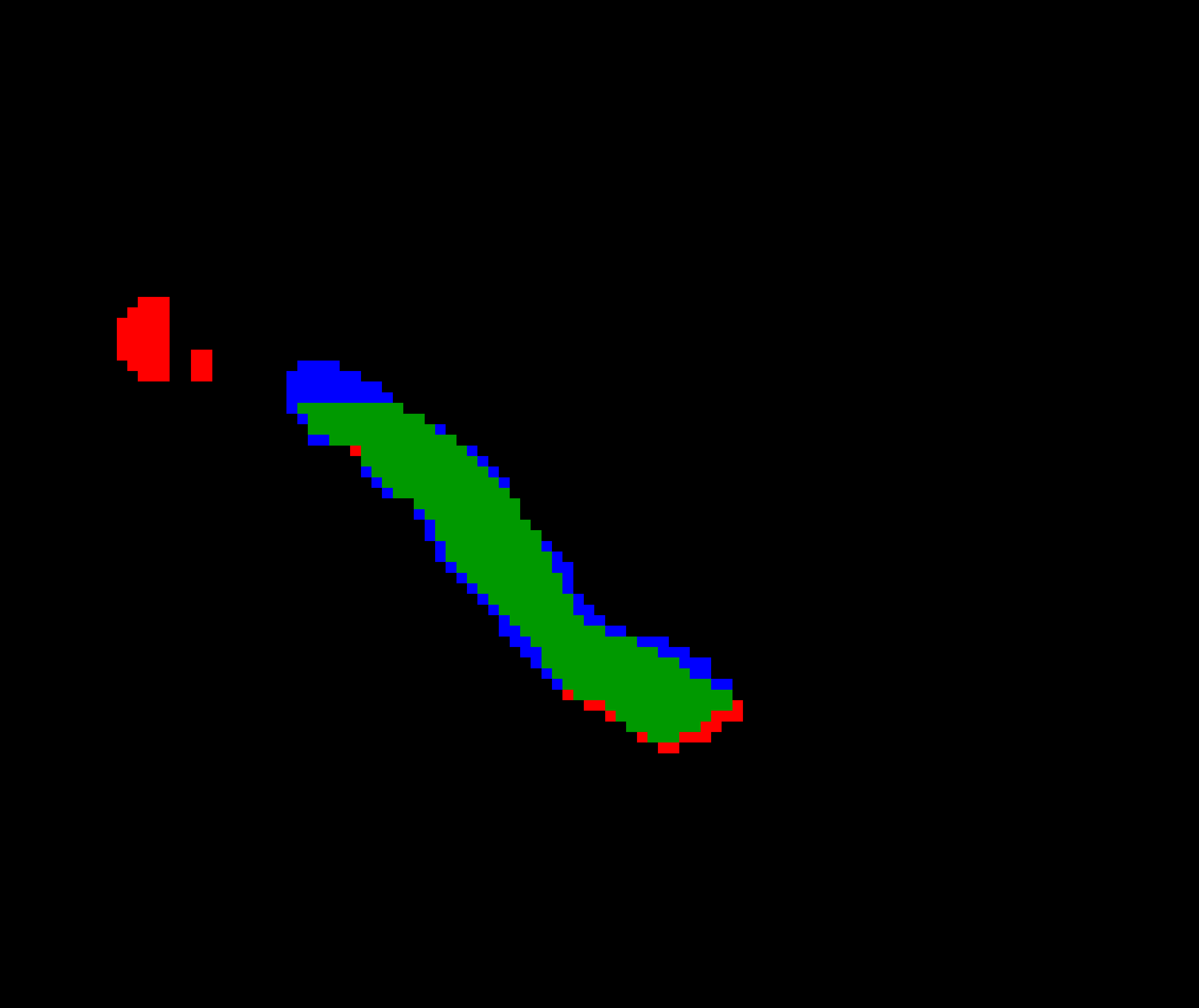} &
\includegraphics[width=2.35cm,height=2.2cm]{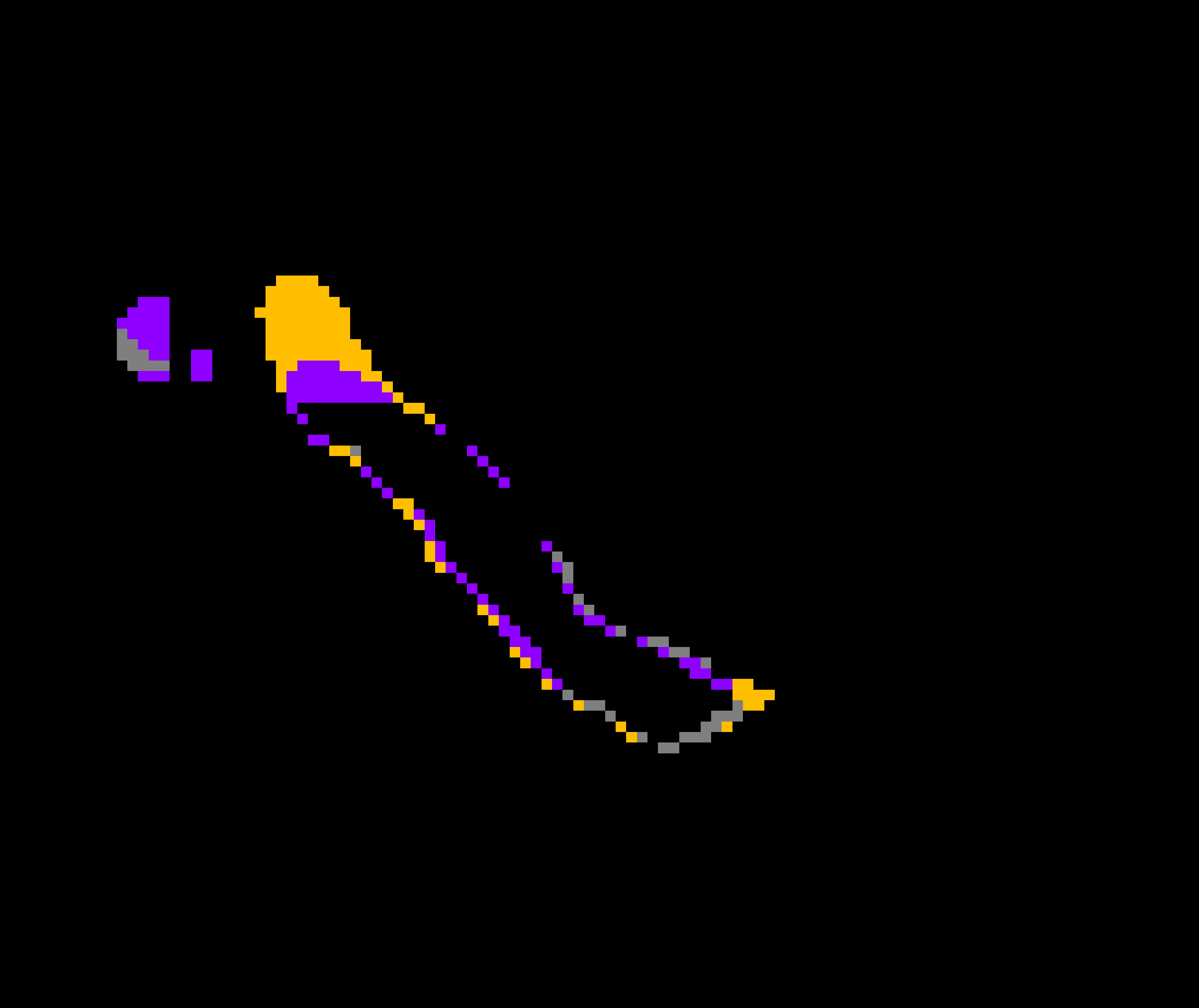} \\
{\scriptsize raw} & {\scriptsize GT} & {\scriptsize base} & {\scriptsize ours} &
{\scriptsize base err.} & {\scriptsize ours err.} & {\scriptsize correction} \\
\end{tabular}
\caption{One row per benchmark (NIH and MSD Task07), body slice. Error overlays colour true positives green, false
positives red and false negatives blue; the correction map marks voxels fixed
by Ours in amber, regressed in violet and wrong in both in grey.}
\label{fig:qual3}
\end{figure*}

\subsection{Comparison with the state of the art}
\label{sec:main}
Table~\ref{tab:panc} compares against published semi-supervised results with one fixed configuration on both datasets. On NIH our model reaches $84.60\%$ DSC, $73.50\%$ Jaccard, $4.97$ voxels 95HD and $1.29$ voxels ASD, i.e.\ $+1.44$ DSC and $-0.47$ 95HD over the strongest compared method (UnCoL \cite{uncol2025}, 83.16/5.44); on MSD it reaches $78.90\%$, $65.85\%$, $4.19$ and $1.19$ ($+2.89$ DSC and $-0.97$ 95HD over DMC \cite{saleknia2026distributional}, 76.01/5.16). The compared methods include the most recent pancreas methods on the same splits, among them DMC (built on TraCoCo) and SDRML (translation consistency with tri-model mutual learning); both remain below our full method on every metric. Gains are larger on distance than on overlap metrics, consistent with removing detached predictions rather than expanding the segmentation. The same configuration works at $19\%$ (NIH) and $50\%$ (MSD) labelled data; PanTS-SSL tests the label budget (Sec.~\ref{sec:pantsres}).

\subsection{PanTS-SSL results}
\label{sec:pantsres}
\begin{table}[!t]
\centering\small
\setlength{\tabcolsep}{4pt}
\caption{PanTS-SSL; labelled budget in parentheses, tumor track with $80$ labelled cases stratified by lesion size.}
\label{tab:pants}
\begin{tabular}{lcccc}
\hline\hline
Configuration & DSC(\%)$\uparrow$ & Jacc(\%)$\uparrow$ & 95HD$\downarrow$ & ASD$\downarrow$ \\
\hline
TraCoCo \cite{liu2024translation} (100) & 83.51 & 73.13 & 10.73 & 2.22 \\
AR+G (50) & 83.57 & 72.85 & 11.10 & 2.72 \\
AR+G (100), seed 2024 & 83.76 & 73.42 & 11.34 & 2.67 \\
AR+G (100) & 84.09 & 73.85 & 9.56 & 2.44 \\
AR+G, tumor track (80) & 84.17 & 73.82 & 9.13 & 2.37 \\
AR+G (250) & 85.29 & 75.49 & \textbf{8.55} & 1.96 \\
AR+G+HAH (100) & \textbf{85.50} & \textbf{75.80} & 9.69 & \textbf{1.27} \\
\hline\hline
\end{tabular}
\end{table}
In Table~\ref{tab:pants}, with 100 labelled volumes AR+G exceeds the co-training baseline on the same budget by $0.58$ Dice ($84.09$ against $83.51$) and reduces 95HD from $10.73$ to $9.56$; with 50 labels it already matches the baseline at 100, and Dice climbs monotonically to $85.29$ at 250 labels while 95HD falls to $8.55$. The tumor track, despite scarce sub-centimetre lesions, attains the best 95HD of the 100-label regime ($9.13$). Seeds at 100 labels differ by $0.33$ Dice; averaged over both, AR+G exceeds the baseline by $0.42$ Dice, while its 95HD gain ($9.56$ and $11.34$ against $10.73$) lies within seed variation. Adding HAH at 100 labels (the full method) raises Dice to $85.50$ and Jaccard to $75.80$ and cuts ASD to $1.27$, at a small 95HD cost ($9.56$ to $9.69$), the same overlap-sharpening trade-off as on NIH.

\subsection{Ablations}
\label{sec:ablations}
\begin{table}[!t]
\centering\small
\setlength{\tabcolsep}{4pt}
\caption{Component ablation on NIH (better network; runs of Table~\ref{tab:panc}); TraCoCo as reported by its authors.}
\label{tab:abl}
\resizebox{\columnwidth}{!}{%
\begin{tabular}{@{}lcccc@{}}
\hline\hline
Configuration & DSC(\%)$\uparrow$ & Jacc(\%)$\uparrow$ & 95HD$\downarrow$ & ASD$\downarrow$ \\
\hline
TraCoCo \cite{liu2024translation} & 81.80 & 69.56 & 5.70 & 1.49 \\
\hline
AR only & 83.32 & 71.72 & 5.41 & 1.35 \\
G only & 84.37 & 73.21 & 4.95 & 1.30 \\
AR $+$ G & 84.39 & 73.26 & \textbf{4.92} & \textbf{1.29} \\
AR $+$ G $+$ HAH & \textbf{84.60} & \textbf{73.50} & 4.97 & \textbf{1.29} \\
\hline\hline
\end{tabular}}
\end{table}

\noindent\textbf{The components address different metric families.} On NIH, Rectification alone brings 95HD down to $5.41$, Grafting alone raises Dice to $84.37$, and together they give $84.39$ Dice with the best 95HD ($4.92$); adding HAH raises Dice to $84.60$ and Jaccard to $73.50$ at a negligible 95HD change ($4.97$): Rectification and Grafting target distinct failure modes, and HAH sharpens overlap. HAH is ablated on top of AR+G, where it is designed to act (Sec.~\ref{sec:hah}); on the three pancreas benchmarks it consistently raises overlap (NIH $84.39$ to $84.60$, MSD $78.61$ to $78.90$, PanTS-SSL $84.09$ to $85.50$ DSC) and raises 95HD by at most $0.22$ while keeping or lowering ASD (PanTS-SSL $2.44$ to $1.27$), since a surrogate averaging a distance-weighted error over the crop acts on mean surface distance rather than on the 95th percentile.

\noindent\textbf{Training-time versus inference-time filtering.} Inference-time filtering is worth little to the starting point: in its published evaluation it moves NIH Dice from $81.80$ to $81.99$ and 95HD from $5.70$ to $5.27$ \cite{liu2024translation}. Our results are all raw network outputs, and the full method reaches the lowest 95HD and ASD of all compared methods on NIH and MSD, so the network trained on rectified and grafted targets produces far fewer detached fragments by itself; even the filtered starting point ($81.99$ DSC, $5.27$ 95HD) stays below our unfiltered full method ($84.60$, $4.97$). A filter removes fragments after the fact; rectified training removes the behaviour that creates them.

\noindent\textbf{What Rectification rejects, and where its effect lives.} Of the components the rule rejects on the baseline's crops, $88.8\%$ on NIH have no ground-truth overlap and are detached hallucinations; the $11.2\%$ overlapping true anatomy lie close to the main component (median $7.0$ voxels) or touch the crop boundary ($56.7\%$), whereas hallucinations sit far away (median $21.7$). On MSD the rejection precision is $95.6\%$, and on thin (tail-like) slices Ours exceeds the baseline by $+0.68$ (NIH) and $+1.33$ (MSD) DSC, so thin anatomy is not eroded. Table~\ref{tab:abl} reports the better network, which hides a like-for-like effect: for the same NIH network (the second), adding Rectification to Grafting improves overlap and distance, from $84.21$ to $84.39$ DSC and $4.99$ to $4.92$ 95HD, so Rectification removes hallucinations that Grafting alone leaves, while Grafting supplies the aggregate overlap gains and HAH sharpens them further.

\noindent\textbf{Coverage and placement.} Doubling graft coverage from the default $23\%$ to $45\%$ degrades every metric for both networks (e.g., 95HD $6.77$ to $10.34$ and $6.09$ to $8.73$; Table~\ref{tab:graft}). Random placement gives similar aggregate metrics, so the gain comes from the transplanted correct annotation, not from placement; surface placement remains the default because it spends the budget where targets are worst (Sec.~\ref{sec:graft}), but we do not claim it raises accuracy: the contribution of G is exact annotation inside the co-training target. Grafting also improves the supervision itself: against the composite reference induced by grafting, the admitted shell accuracy of the target rises on both datasets, most where the peer target is weakest, the mechanism behind the scaling in Table~\ref{tab:pants}.

\begin{table}[!t]
\centering\small
\setlength{\tabcolsep}{4pt}
\caption{Graft design on NIH, Rectification disabled, both networks. Doubling coverage from $23\%$ to $45\%$ degrades every metric; surface and random placement are near-equivalent, so the benefit comes from the transplanted annotation.}
\label{tab:graft}
\begin{tabular}{llcccc}
\hline\hline
Setting & Net & DSC$\uparrow$ & Jaccard$\uparrow$ & 95HD$\downarrow$ & ASD$\downarrow$ \\
\hline
\multicolumn{6}{l}{\emph{Coverage}} \\
$23\%$ & 1 & 83.73 & 72.26 & 6.77 & 2.41 \\
       & 2 & 83.80 & 72.40 & 6.09 & 2.10 \\
$45\%$ & 1 & 83.39 & 71.80 & 10.34 & 2.75 \\
       & 2 & 83.29 & 71.67 & 8.73 & 2.78 \\
\hline
\multicolumn{6}{l}{\emph{Placement at $23\%$ coverage}} \\
Surface & 1 & 83.73 & 72.26 & 6.77 & 2.41 \\
        & 2 & 83.80 & 72.40 & 6.09 & 2.10 \\
Random & 1 & 83.41 & 71.84 & 6.18 & 2.08 \\
        & 2 & 83.92 & 72.53 & 7.58 & 2.23 \\
\hline\hline
\end{tabular}
\end{table}

\subsection{Generalization Beyond the Pancreas}
\label{sec:general}
To test transfer to another single-component anatomy and a second modality, we apply the identical fixed configuration, with no per-dataset tuning, to the Left Atrium (LA) benchmark with the splits and protocols of \cite{liu2024translation}. The LA cavity is a single connected structure, so the connectivity prior of Sec.~\ref{sec:ar} applies directly, and MRI adds a CT-to-MRI shift. With 16 labeled volumes (20\%), Table~\ref{tab:sgeneral} shows that Ours improves every metric over the published baseline and two recent methods \cite{xie2026spatially,lu2025enhancing}, most on 95HD and ASD, consistent with suppressing confident detached fragments.

\begin{table}[!htb]
\centering\small
\setlength{\tabcolsep}{4pt}
\caption{Generalization to the Left Atrium benchmark (16 labeled).}
\label{tab:sgeneral}
\resizebox{\columnwidth}{!}{%
\begin{tabular}{@{}llcccc@{}}
\hline\hline
Dataset & Method & DSC(\%)$\uparrow$ & Jacc(\%)$\uparrow$ & 95HD$\downarrow$ & ASD$\downarrow$ \\
\hline

LA
& TraCoCo~\cite{liu2024translation} & 91.51 & 84.40 & 5.63 & 1.79 \\

& DTC \cite{luo2021semi} & 89.42&80.98&7.32&2.10 \\
& MC-Net+ \cite{wu2022mutual} & 90.34&	82.48&	6.00& 1.77\\
& SDRML \cite{xie2026spatially} & 90.22 & 82.18 & 6.13 & 1.79 \\
& CCE \cite{lu2025enhancing} & 90.90 & 83.41 & 5.28 & 1.63 \\
& Ours & \textbf{92.07} & \textbf{85.38} & \textbf{5.21} & \textbf{1.32} \\
\hline\hline
\end{tabular}}
\end{table}

\subsection{Qualitative analysis}
\label{sec:qual}
\begin{figure}[!t]
\centering\small
\setlength{\tabcolsep}{2pt}
\begin{tabular}{@{}ccc@{}}
& {\scriptsize base} & {\scriptsize Ours} \\
{\scriptsize NIH} &
\includegraphics[width=3.9cm,height=2.9cm,keepaspectratio]{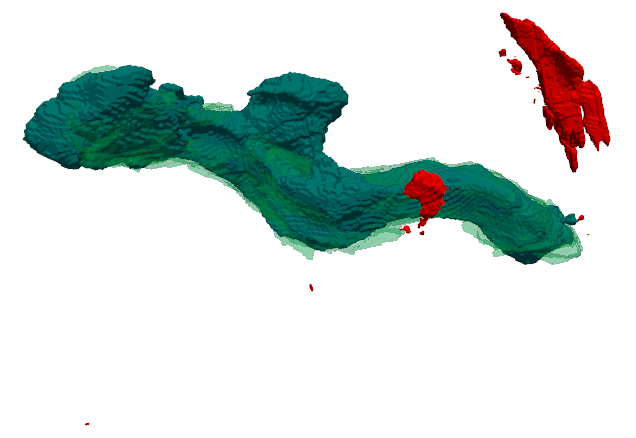} &
\includegraphics[width=3.9cm,height=2.9cm,keepaspectratio]{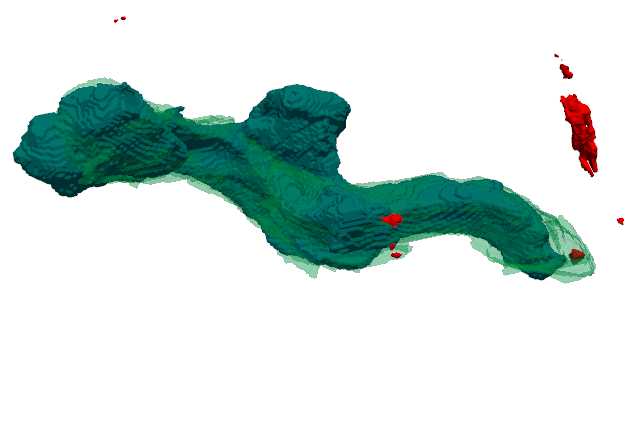} \\
{\scriptsize MSD} &
\includegraphics[width=3.9cm,height=2.9cm,keepaspectratio]{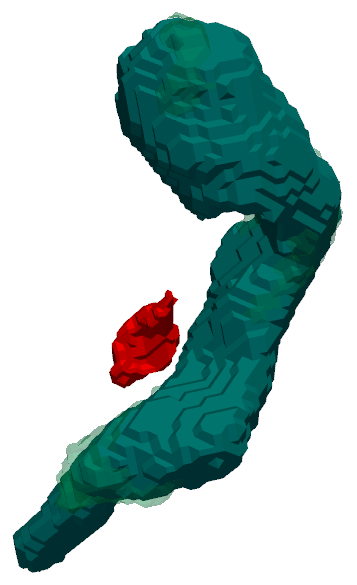} &
\includegraphics[width=3.9cm,height=2.9cm,keepaspectratio]{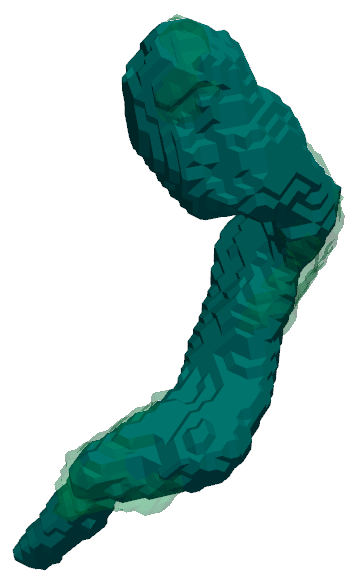} \\
{\scriptsize PanTS} &
\includegraphics[width=3.9cm,height=2.9cm,keepaspectratio]{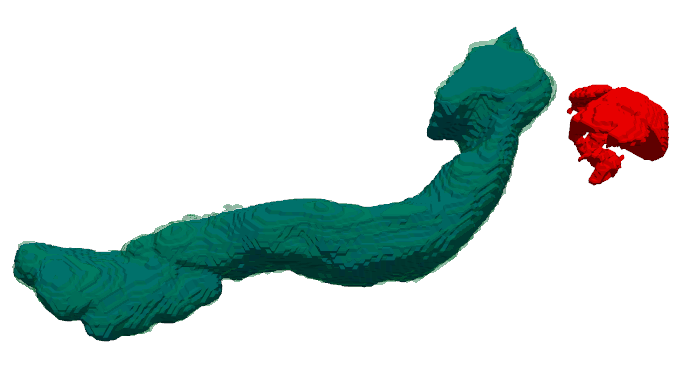} &
\includegraphics[width=3.9cm,height=2.9cm,keepaspectratio]{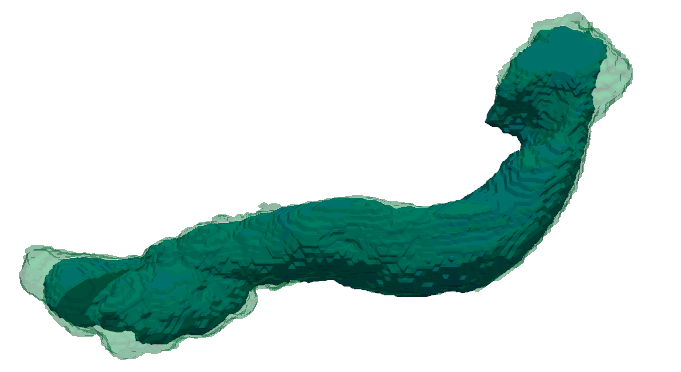} \\
\end{tabular}
\caption{3D surfaces per benchmark; ground truth translucent green, main component teal, detached fragments red. Baseline meshes carry hallucinated fragments far from the organ; Ours removes them.}
\label{fig:3d}
\end{figure}
Figs.~\ref{fig:evol} to~\ref{fig:3d} show the failure modes: in every row of Fig.~\ref{fig:qual3} the baseline error map has large red islands far from the organ, the signature of confident hallucination, whereas Ours is almost entirely green along the shell and its correction maps are dominated by amber with negligible violet, so Ours fixes errors and rarely regresses. Fig.~\ref{fig:3d} confirms that the 95HD reductions in Tables~\ref{tab:panc} and~\ref{tab:pants} are fragment removals, not mild smoothing, and Fig.~\ref{fig:evol} shows fragments fluctuating but shrinking, with one small fragment left at the end, as rectified targets teach where the organ cannot be and grafted voxels keep the shell improving.

\section{Conclusion}
\label{sec:conclusion}
The pancreas is a single connected structure, so confident foreground fragments detached from the organ are anatomically impossible, yet confidence gates are blind to them because the errors themselves are confident. We therefore looked outside the prediction: anatomy supplies the criterion (Anatomical Rectification), annotation the targets (Boundary Grafting), and geometry the penalty (Homotopy-Annealed Hausdorff Supervision); AR and G add no loss term and leave the targets unchanged when their trigger is absent. With one fixed configuration, the method improves all four metrics on NIH and MSD, reaching $84.60\%$ Dice and $4.97$ voxels 95HD, transfers to the Left Atrium MRI benchmark without per-dataset tuning, and converts unlabelled data into supervision that scales monotonically on PanTS-SSL. For targets that form one connected structure, two lessons follow: a plausibility criterion drawn from structure reaches errors that confidence and uncertainty cannot rank, and where a prior is applied matters, since acting on the training target changes what the network learns while cleaning the output leaves the suppressed behaviour inside the model. AR assumes a single connected target; organs with several valid components or multi-class tasks would need a per-class component count.

{
    \small
    \bibliographystyle{ieeenat_fullname}
    \bibliography{main}
}

\end{document}